\documentclass[pdflatex,sn-mathphys-num]{sn-jnl}% Math and Physical Sciences Numbered Reference 

\usepackage{graphicx}%
\usepackage{amsmath,amssymb,amsfonts}%
\usepackage{amsthm}%
\usepackage[mathscr]{euscript}
\usepackage[boxed,lined,linesnumbered, english]{algorithm2e}
\usepackage{subfigure}
\usepackage{tikz}
\usetikzlibrary{arrows,petri,topaths,shapes,trees,positioning,backgrounds}
\tikzstyle{line} = [draw, -latex']
\usepackage{afterpage}      % Para \afterpage
\usepackage{pdflscape}      % Para \begin{landscape} ... \end{landscape}
\usepackage{silence}
\usepackage[labelfont=bf]{caption} % sem hypcap global
\usepackage[all]{hypcap}            % só ativa onde funciona
\usepackage{makecell}       % Para \setcellgapes e \makegapedcells
\usepackage{multirow}
\usepackage[numbers,sort&compress]{natbib}%
\newcommand{\uset}[1]{\ifmmode\left\{\,#1\,\right\}\else\{\,#1\,\}\fi}

\theoremstyle{thmstyleone}%
\theoremstyle{thmstyletwo}%

\theoremstyle{thmstylethree}%
\newtheorem{definition}{Definition}%

\begin{document}

\title[Impact of diversity on bounded archives for multi-objective
local search]{Impact of diversity on bounded archives for multi-objective
local search}


\author*[1]{\fnm{Amadeu} \sur{A. Coco}}\email{amadeu.almeidacoco@univ-lille.fr}

\author[2]{\fnm{Cyprien \sur{Borée}}}\email{cyprien.boree@tuta.io}

\author[1]{\fnm{Julien} \sur{Baste}}\email{julien.baste@univ-lille.fr}

\author[1]{\fnm{Laetitia} \sur{Jourdan}}\email{laetitia.jourdan@univ-lille.fr}
\author[1,2]{\fnm{Lucien} \sur{Mousin}}\email{lucien.mousin@univ-catholille.fr}
\affil*[1]{\orgdiv{ORKAD, CRIStAL, UMR 9189, CNRS, Centrale Lille}, \orgname{Université de Lille}, \orgaddress{\city{Lille}, \country{France}}}

\affil[2]{\orgdiv{FGES}, \orgname{Lille Catholic University, F-59000 , France}, \orgaddress{\city{Lille}, \country{France}}}


\abstract{This works tackles two critical challenges related to the development of metaheuristics for Multi-Objective Optimization Problems (MOOPs): the exponential growth of non-dominated solutions and the tendency of metaheuristics to disproportionately concentrate their search on a subset of the Pareto Front. To counteract the first, bounded archives are employed as a strategic mechanism for effectively managing the increasing number of non-dominated solutions. Addressing the second challenge involves an in-depth exploration of solution diversity algorithms found in existing literature. Upon recognizing that current approaches predominantly center on diversity within the objective space, this research introduces innovative methods specifically designed to enhance diversity in the solution space. Results demonstrate the efficacy of the Hamming Distance Archiving Algorithm, one of the newly proposed algorithms for multi-objective local search, surpassing the performance of the Adaptive Grid Archiving and the Hypervolume Archiving, both drawn from the literature. This outcome suggests a promising avenue for enhancing the overall efficiency of metaheuristics employed for solving MOOPs.}


\maketitle

\section{Introduction}\label{s:introdution}

Metaheuristics represent a widely adopted strategy for addressing Multi-Objective Optimization Problems (MOOPs)~\cite{blot2018}. Typically applied in scenarios where traditional optimization techniques are proven impractical~\cite{blot2018}, these approaches have witnessed a substantial surge in development recently~\cite{legrand2023,segura2023}. Consequently, various effective and versatile algorithms have emerged. However, despite their success, some challenges persist, impacting their overall performance. One notable issue is the exponential growth in the number of non-dominated solutions identified by these algorithms. Additionally, there is a tendency for metaheuristics to focus their search disproportionately on a subset of the Pareto Front, resulting in challenges related to solution diversity.

This study addresses two critical challenges mentioned in the previous paragraph. To deal with the exponential growth, the use of bounded archives~\cite{laumanns2002,segura2023} is employed as a strategic mechanism for managing the increasing number of non-dominated solutions. As for the second challenge, an extensive examination of solution diversity algorithms presented in the literature~\cite{knowles2002,knowles2003} is undertaken. After noting that the existing approaches primarily focus on diversity within the objective space, this work introduces novel methods that address diversity in the solution space, aiming to demonstrate the efficacy of this avenue for enhancing the overall efficiency of metaheuristics tailored for solving MOOPs.

Over the course of the past forty years, numerous studies~\cite{schaffer2014,hansen1997,czyzzak1998,deb2000,knowles2000,talbi2001,zhang2007,geiger2008,liefooghe2012} have introduced a diverse range of metaheuristics for MOOPs. The authors of the survey~\cite{blot2018} have categorized these algorithms into three distinct groups, which will be referred to as Algorithmic Families (AF) from this point forward. Each AF consists of a collection of diverse approaches that share a common origin, such as evolutionary methods, generalization of single-objective metaheuristics, and Pareto local search algorithms specifically proposed for MOOPs.

The algorithmic family of evolutionary methods is represented by MOOP-solving techniques based on mainly genetic algorithms~\cite{schaffer2014,deb2000,zhang2007}. According to our knowledge, the first developed metaheuristic for MOOPs is the Vector Evaluated Genetic Algorithm~\cite{schaffer2014}. Other well-known evolutionary methods that deal with MOOPs are the Nondominated Sorting Genetic Algorithm II~\cite{deb2000} and the Multiobjective Evolutionary Algorithm Based on Decomposition~\cite{zhang2007}. To the reader interested in a more detailed description of evolutionary algorithms for MOOPs, the survey~\cite{blot2018} is recommended.

The Single Objective Metaheuristics Generalizations family encompasses algorithms that have evolved from metaheuristics originally designed for single-objective optimization problems~\cite{hansen1997,czyzzak1998,geiger2008}. Notable adaptations within this family include the Multi-Objective Tabu Search~\cite{hansen1997}, the multi-objective Simulated Annealing~\cite{czyzzak1998}, and the multi-objective variable Neighborhood Search~\cite{geiger2008}. For a comprehensive exploration of metaheuristics for MOOPs rooted in single-objective methods, readers are directed to reference~\cite{amine2019}.

The Pareto Local Search (PLS) Algorithmic Family~\cite{knowles2000,talbi2001,liefooghe2012} is a significant category in MOOPs metaheuristics, focusing on Pareto Local Optima (PLO) to refine the Pareto Front. PLO encompasses a set of solutions with no dominating neighbors, where each neighbor is either dominated or incomparable. It is essential to note that while a Pareto optima is always a PLO, not all PLO are guaranteed to be Pareto optima, regardless of the neighborhood. Notable PLS metaheuristics include Pareto Archived Evolution Strategy~\cite{knowles2000}, Pareto Local Search 2~\cite{talbi2001}, and Dominance-based multi-objective local search~\cite{liefooghe2012}. For a comprehensive exploration of PLS-based approaches in MOOPs, readers are referred to the survey in~\cite{blot2018}.

The reminder of this manuscript is organized as follows. Section~\ref{s:definition} introduces the key concepts regarding multi-objective optimization, archive and solution diversity. Section \ref{s:obj-space} presents archiving algorithms developed in the literature and their diversity issues. Subsequently, the new approaches for solution diversity algorithms in bounded archives are detailed in Section~\ref{s:diversity}. These methods are evaluated using a set of computational experiments described in Section~\ref{s:experiments}. Finally, concluding remarks and perspectives are discussed in Section~\ref{s:conclusion}.

\section{Definitions}\label{s:definition}

A multi-objective optimization problem is described by at least two objective functions in~\eqref{eq:mo-objectives} and the constraints in~\eqref{eq:mo-constraints}. Without loss of generality, it is assumed that all objective functions are minimization ones. In this generic problem, the set of objective functions is denoted as $f = \{f^1, f^2, \ldots, f^m\}$, as outlined in Equation~\eqref{eq:mo-objectives}. The inequalities presented in~\eqref{eq:mo-constraints} represent the constraints of a problem, resulting in the formation of a polytope $\mathscr{P}$. Additionally, the set of feasible solutions within $\mathscr{P}$ is represented by $\Phi$, and a feasible solution is denoted as $\phi \in \Phi$.

It is worth noting that the function $f: \mathbb{R}^{n} \rightarrow \mathbb{R}^{m}$ maps solutions within $\Phi$ into a $m$-dimensional space of solution images denoted as $\Psi$. This mapping is expressed as $\Psi = { \psi \in \mathbb{R}^{m}: \psi = f(\phi), \phi \in \Phi}$. Therefore, each position $i$ in vector $\psi$ corresponds to the value of the objective function $f^i$ for a given solution $\phi$, and it is indicated as $\psi_i = f^i(\phi)$.

\begin{align}
    & \min\limits_{\phi \in \Phi} \quad [f^1(\phi), f^2(\phi), \ldots, f^m(\phi)]^T \label{eq:mo-objectives} \\ 
    & s.t. \nonumber \\
    & \quad \quad \quad q_j(\phi) \leqslant 0,~j = 1, \ldots, J. \label{eq:mo-constraints}
\end{align}

The five subsequent definitions, as referenced from Deb~\cite{Deb2014}, characterize the \textit{Pareto-front} of a multi-objective optimization problem.

\begin{definition}
A solution $\phi_1 \in \Phi$ \emph{dominates} another solution $\phi_2 \in \Phi$, if and only if, $f^i(\phi_1) \leqslant f^i(\phi_2)$ for all objective functions $\uset{f^1, f^2, \ldots, f^m}$ and $f^i(\phi_1) < f^i(\phi_2)$ for at least one objective function $f^i$. Henceforth, this relation is denoted as $\phi_1 \prec \phi_2$
\end{definition}

\begin{definition}
A solution $\phi' \in \Phi$ is said to be \emph{non-dominated} if and only if there is no solution $\phi \in \Phi$ such that $\phi \prec \phi'$.
\end{definition}

\begin{definition}
A point $\psi^* \in \Psi$ is said to be \emph{Pareto-optimal} if and only if there is no point $\psi \in \Psi$ such that $\psi_i \leqslant \psi^*_i$, for $i \in \{1, \ldots, m\}$, and that $\psi_i < \psi^*_i$, for at least one $i \in \{1, \ldots, m\}$.
\end{definition}

\begin{definition}
The \emph{Pareto-front} is the set of the Pareto-optimal points.
\end{definition}

\begin{definition}
 The \emph{nadir point} is a solution $\phi_{np} \in \Phi$ that contains the worst possible values for each objective among all solutions that belong to the Pareto-front.
\end{definition}

\begin{figure}[ht]
\centering \includegraphics[width=.9\linewidth]{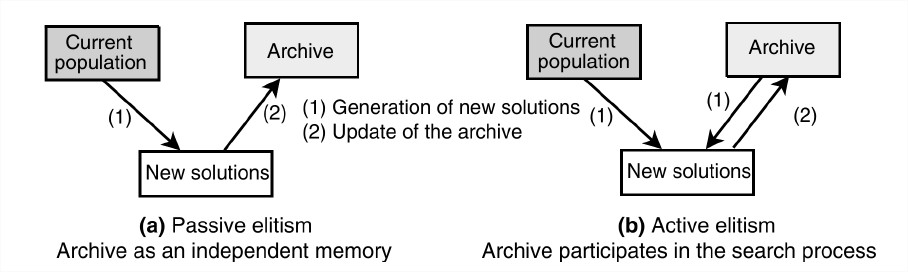}
\caption{Cases of archive utilization. \label{archives}}
\end{figure}

The concepts of archive and solution diversity are pivotal in multi-objective optimization. An \textbf{archive} is a data structure that stores and manages a set of non-dominated solutions found during the performance of an MOOP metaheuristic. It is commonly used to maintain a diverse and representative set of solutions, helping in the convergence analysis and in the exploration and improvement of the Pareto front. An archive may be bounded, storing a fixed number of solutions, or dynamically managed. However, the number of solutions of a dynamically managed archive can grow exponentially, often resulting in algorithmic slowdowns due to the exploration of a large solution space by a MOOP metaheuristic \cite{liefooghe2012}. Moreover, as illustrated in Figure~\ref{archives}, an archive can be either passive or active. In the former, it works as a secondary population that solely stores non-dominated solutions. In the latter, it serves as a base population that guides the metaheuristic search. This study exclusively focuses on bounded active archives, which play a critical role in many multi-objective optimization approaches. They ensure the discovery of a diverse and representative set of solutions while addressing the challenges posed by the exponential increase in non-dominated solutions \cite{knowles2003}.

To better represent the Pareto Front of a MOOP, the solutions within an archive must exhibit a high degree of diversity. Thus, the \textbf{solution diversity} serves as a pivotal metric, providing insight into the range and dissimilarity of the solutions stored in the archive. A diverse archive presents an opportunity for a comprehensive exploration of the solution space. In the pursuit of enhancing MOOP metaheuristics, the primary objective of this work lies in developing novel approaches that effectively improve solution diversity within bounded archives. These approaches will be detailed in Section~\ref{s:diversity}.

\section{Objective space archiving algorithms} \label{s:obj-space}

In this section, the exploration of diversity algorithms based on the objective space is presented. Subsection~\ref{ss:aga} provides a description of the Adaptive Grid Archiving. Then, the Hypervolume Archiving method is introduced in Subsection~\ref{ss:ha}. After that, diversity issues within objective space algorithms are addressed in Subsection~\ref{ss:issues}.

\subsection{Adaptive Grid Archiving}
\label{ss:aga}

The Adaptive Grid Archiving (AGA)~\cite{knowles2002} introduces an innovative approach to deal with solution archives in multi-objective optimization. This method employs a grid-based framework to effectively partition the objective space, facilitating the categorization of solutions into distinct spatial regions. As the archive nears its capacity, AGA works as follows. First, it prioritizes the preservation of non-dominated solutions originating from less congested areas. Conversely, solutions from the most densely populated regions are randomly and uniformly removed. This method proves instrumental in maintaining well-distributed solutions along the Pareto front, enhancing the diversity and quality of the solution set. 

The division of the objective space is given by a predefined parameter, which we will refer to as $div$. This parameter plays a critical role in shaping how the solution diversity management process operates. The set of solutions that can be removed from the archive is directly influenced by the value of $div$. If the value of $div$ is almost as large as the archive's capacity, it may lead to a scenario where each grid region contains at most one solution. On the other hand, if the value of $div$ is too small, a crowded archive with an excessive number of solutions in each region can be obtained. Both scenarios have the potential to compromise the effectiveness of approximating the Pareto front. Determining an optimal value for $div$ is complex, as it depends on both the archive's capacity and the number of objectives. Given archive capacity $AC$ and the number of objectives $o$, it was demonstrated by~\cite{knowles2002} that the convergence is guaranteed if the inequality~\eqref{AGA1} is satisfied.

\begin{equation}
AC > div^o - (div - 1)^o + 2 \times k
\label{AGA1}
\end{equation} 

Expressing archive capacity $AC$ in terms of $o$ and $div$ is straightforward, as outlined in inequality~\eqref{AGA1}. However, when dealing with limited memory and consequently restricted archive capacity, it becomes necessary to express div as a function of both $o$ and $AC$, as indicated by equation~\eqref{AGA2}. Finding a closed-form expression for this inequality that holds true for any $o$ is a challenging endeavor. One approach to overcome this challenge is to iteratively test various values of $div$ until the inequality is satisfied. 

\begin{equation}
AC - 2 \times o > div^o - (div - 1)^o 
\label{AGA2}
\end{equation} 

\begin{figure}[!htbp]
\centering  

\subfigure[]  
{  
    \centering \includegraphics[scale=0.25]{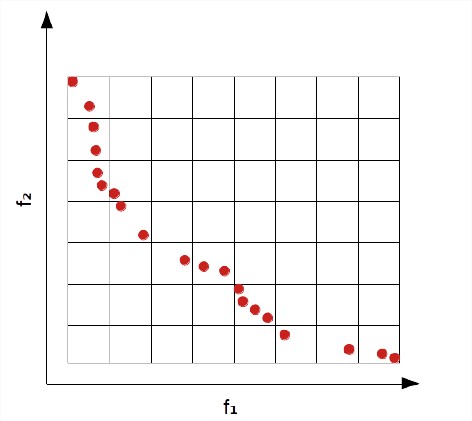}
} 
\subfigure[]  
{  
    \centering \includegraphics[scale=0.25]{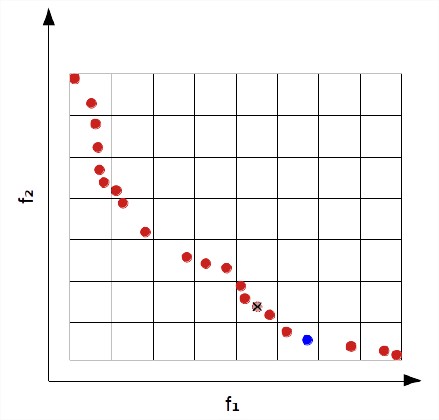}
}
\caption{(a) initial Pareto Front; (b) Pareto Front updated by AGA.}
\label{fig:AGA}
\end{figure}

In Figure~\ref{fig:AGA}, an example of a bounded archive capable of holding up to 20 solutions is presented, as represented by the distinct red dots in Figure~\ref{fig:AGA}(a). To effectively manage these solutions, $div$ is set to $8$, serving as the foundation for establishing a grid system based on objective extremums. This grid is a crucial tool for monitoring the population within each grid region, aiding in the identification of crowded areas. When a new non-dominated solution is generated, as depicted by the blue dot in Figure~\ref{fig:AGA}(b), the AGA algorithm ensures that the archive remains well-maintained. To achieve this, the following approach is employed. When a new non-dominated solution is found, it displaces a randomly selected solution from an overcrowded area, thus maintaining a balanced and diverse representation of solutions in the archive.

\subsection{Hypervolume Archiving}
\label{ss:ha}

\begin{figure}[ht]
\centering \includegraphics[width=.5\linewidth]{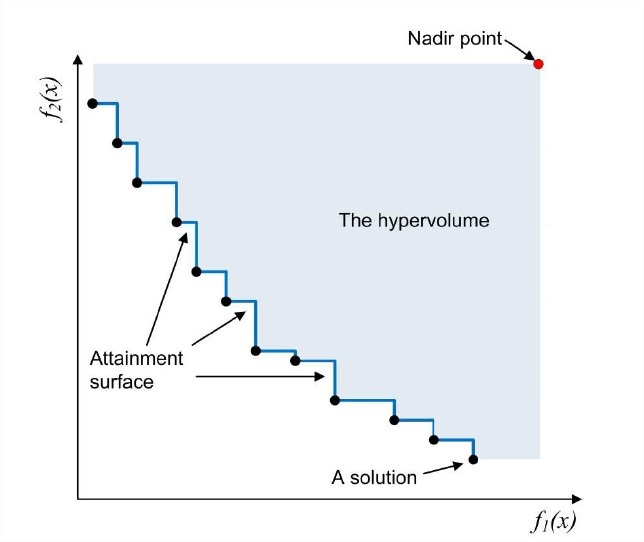}
\caption{Example of hypervolume on two objectives, including the nadir point.\label{HA1}}
\end{figure}

The hypervolume, also known as the Lebesgue measure, was initially introduced by~\cite{zitzler1999} as a method for evaluating the outcome of metaheuristics in multi-objective optimization. In the study conducted in~\cite{knowles2003}, this metric was adapted for the purpose of bounding archives. When an archive reaches its full capacity, a new non-dominated solution is admitted if and only if it contributes to increasing the hypervolume more than some other solution already there. The objective of the hypervolume is to compute the volume occupied between the non-dominated solutions in the archive and the nadir point, as illustrated in Figure~\ref{HA1}. This approach aids in assessing the diversity and coverage of solutions within the archive in multi-objective optimization.

In two and three dimensions, the hypervolume corresponds to the area and volume dominated by the solution-set, respectively. However, in higher dimensions, the dominated volume is represented as a convex $n$-polytope. It is worth noting that as the dimensionality increases, the computational complexity of calculating the hypervolume also rises. When the hypervolume is employed as a one-time evaluation of the final archive quality, a straightforward algorithm is usually sufficient. However, during an archiving process, it is critical to minimize computational complexity, as each addition or removal of a single solution directly impacts the entire dominated volume. Given that this operation is repeated numerous times during the optimization process, an inefficient implementation can significantly slow down the overall search algorithm. Hence, optimizing the efficiency of the archiving process is of utmost importance.

In~\cite{knowles2003}, the authors employ the Hypervolume Archiving (HA) algorithm, initially developed in~\cite{fleischer2003}, to efficiently calculate changes in the hypervolume when adding or removing a point from an archive. Originally designed for computing the Lebesgue measure of regions dominated by non-dominated points, this algorithm works by identifying and removing non-overlapping dominated regions within a set of non-dominated points. These dominated regions are defined by rectangular polytopes, simplifying the computation of each hypervolume. Consequently, by iteratively eliminating such regions until none remain and aggregating their hypervolumes, it is possible to determine the hypervolume dominated by a set of non-dominated points.

In the HA algorithm, the non-dominated points are placed on a last-in, first-out stack. When the point at the top of the stack is popped, it generates $k$ other points, where $k$ is the number of objectives. Each of these newly spawned points is inserted into the stack if and only if it is not dominated by any point in the stack and if it does not share a component equal to the corresponding component of the bounding point, which is usually the nadir point. The algorithm proceeds by removing the next point from the stack. At each removal, the algorithm computes the volume of the rectangular polytope that is being removed and adds it to a cumulative hypervolume.

According to~\cite{fleischer2003}, the total number of points in the stack cannot exceed $size + k - 1$, where the constant $size$ is the stack's initial depth. This constraint comes from the spawn points, as all but one are dominated by the remaining contents of the stack. Consequently, while spawned points continue to generate further points, the total number of points in the stack remains tightly bounded. This results in a worst-case algorithmic complexity of $\mathscr{O}{(size^3k^2)}$.

If the algorithm is run until the stack is empty, the computation of the hypervolume of entire non-dominated points is performed. However, for archiving purposes, it is enough to calculate the hypervolume contribution of a single point to the set, which is also possible with HA. To determine the contribution of a new non-dominated point, one can simply add it to the stack and execute HA. In this case, the algorithm stops when the stack depth returns to its initial value, indicating the removal of the new point and its spawns. For a more comprehensive description of this procedure, readers are referred to~\cite{zitzler1999}.

\subsection{Diversity issues on objective space algorithms}
\label{ss:issues}

Upon a comprehensive examination of AGA, HA, and similar approaches in the literature~\cite{lopez2011}, two key insights are highlighted. The initial aspect is related to the archive's capacity. Once a bounded archive reaches its limit, it includes a new solution $X'$ only if $X'$ offers a greater contribution according to a diversity criterion compared to another solution already within the archive. This approach is valuable for summarizing the archiving procedure as part of the effort to optimize solution diversity within bounded archives within an algorithmic framework. Secondly, it is worth noting that the majority of archiving methods exclusively operate within the objective space, often neglecting the solution space when addressing solution diversity. When the primary focus of an approach is finding a high-quality approximation of the Pareto front and ensuring well-distributed solutions in terms of fitness, the solution space is usually omitted in favor of the decision space.

On the other hand, addressing solution space diversity is a key point in various MOOP. For instance, in cutting-stock problems~\cite{cherri2014}, a piece of material of standard size and a prescribed set of shapes are provided. The objective of a cutting-stock problem is often to cut the material into pieces of the specified shapes in a way that minimizes the amount of leftover material. In this context, a solution is given by a minimum-size leftover. The availability of a set of suitably diverse solutions provides an opportunity to select an appropriate leftover that may be utilized later in the fabrication of pieces with unspecified shapes. Consequently, the development of archiving algorithms rooted in the solution space emerges as an essential tool for dealing with numerous MOOP challenges.

\section{Approaches for improving solution diversity} \label{s:diversity}

In this section, a novel algorithm developed to improve the solution diversity on bounded archives is presented. Subsection~\ref{ss:solspace} introduces the diversity within the solution space. In the sequel, the approach based on the solution space is described in Section~\ref{ss:sol-space}.

\subsection{Diversity in the solution space}
\label{ss:solspace}

In Section~\ref{s:obj-space}, the concept of solution diversity was introduced as a criterion for the selection of non-dominated solutions kept in a bounded archive. Thus, greater diversity is achieved when, on average, neighboring solutions are as far apart as possible. On the other hand, diversity in the solution space should be measured using another metric, henceforth referred to as the solution metric. The solution metric is defined as a function that quantifies the distance between any two non-dominated solutions within the set. When considering two solutions, $A$ and $B$, the following properties must be satisfied by the metric space:

\begin{enumerate}
    \item The distance from A to B is zero if, and only if, A and B are identical solutions.
    \item The distance between two distinct solutions is always positive.
    \item The distance from A to B equals the distance from B to A.
    \item The distance from A to B is never greater than the distance from A to B via any third solution C.
\end{enumerate} 

The four properties outlined above correspond to key characteristics of a solution metric. These properties respectively define identity of (i) indiscernibility; (ii) non-negativity; (iii) symmetry; and (iv) the triangle inequality. It is important to note that a solution metric function $d$ is represented as $d : S \times S \rightarrow R$, where $S$ represents the solution space. This signifies that the metric's definition is often dependent on a solution representation, which can vary between different MOOP. Consequently, presenting universally applicable metrics can be a challenging task.

\subsection{Solution space archiving algorithm} \label{ss:sol-space}

As detailed in section~\ref{s:definition}, setting a bounded archive is critical to managing the vast number of solutions generated during the metaheuristic search, especially as instance sizes increase. Merely setting a constant as an acceptance criterion for new solutions is usually insufficient. Therefore, three criteria are employed to determine solution acceptance in bounded archives. Firstly, a solution is accepted if it dominates other solutions in the archive. In such cases, all dominated solutions are removed, and the new solution is kept. Secondly, a solution is accepted if it neither dominates other solutions in the archive nor is dominated by them, and the archive is not full. In this scenario, the solution is added to the archive. Lastly, when the archive is full, a new non-dominated solution is accepted only if it contributes more to the bounded archive, following a solution-space diversity metric, than another solution in it. 

In this research, novel solution-space diversity metrics based on the Hamming and Jaccard distances are introduced. These metrics assess the contribution of a non-dominated solution $S$ to the archive. In both cases, the contribution is given by the sum of the Hamming or Jaccard distances between $S$ and all other solutions in the archive, capturing the dissimilarity or overlap between $S$ and the archive solutions. Consequently, the solutions that contribute the most to the archive are those with the highest sum of distances. From now on, these methods are referred to as Hamming Distance Archiving Algorithm (HDAA) and Jaccard Distance Archiving Algorithm (JDAA). This approach to solution-space diversity measurement adds a nuanced perspective to the evaluation and selection of non-dominated solutions, enhancing the diversity in bounded archives.

Hamming distance is widely recognized in the literature for its significant applications in error detection and correction within coding theory. It measures the dissimilarity between two equal-length strings of symbols by counting the positions at which their corresponding symbols differ. For instance, if solutions of a  MOOP are represented using binary vectors, the Hamming distance $S1 \bigoplus S2$ between $S1 = 111100110101$ and $S2 = 000000110100$ is computed as the number of differing 'ones'. Thus, $S1 \bigoplus S2 = 6$

The Jaccard index (JI) is a widely used statistical measure for assessing the similarity and dissimilarity of sample sets. It is defined as the size of the intersection divided by the size of the union of two distinct sample sets, as indicated in Equation~\eqref{jaccindex}. Given two sample sets $A$ and $B$, it is important to note that $0 \leq JI(A,B) \leq 1$. When both sets $A$ and $B$ are empty, $J(A, B) = 1$. The Jaccard index is commonly applied in fields using binary data, such as computer science, ecology, and genomics. In contrast, the Jaccard distance (JD), which quantifies the dissimilarity between sample sets, is the complement of $JI$, meaning $JD(A,B) = 1 - JI(A,B)$. This calculation is detailed in equation~\eqref{jaccdistance}.

\begin{equation}
J(A,B) = \frac{ \vert A \cap B \vert}{\vert A \cup B \vert} = \frac{\vert A \cap B \vert}{\vert A \vert + \vert B \vert - \vert A \cap B \vert}
\label{jaccindex}
\end{equation} 

\begin{equation}
d_J(A,B) = 1 - J(A,B) = \frac{\vert A \cup B \vert - \vert A \cap B \vert}{\vert A \cup B \vert}
\label{jaccdistance}
\end{equation} 

\vspace{0.4cm}

Algorithm \ref{alg:1} provides a detailed description of HDAA. Given a non-dominated solution $\phi$ found by a MOOP metaheuristic and an archive $\alpha_{old}$ as inputs, the algorithm produces an updated archive $\alpha_{new}$ as its output. The iterative processing of each solution in $\alpha_{old}$ takes place within the loop in lines $1-5$. In each iteration, if any solution $\phi_i \in \alpha_{old}$ is dominated by $\phi$, it is removed from $\alpha_{old}$ in line 3. In line 6, a new archive $\alpha_{new}$ is created by merging $\alpha_{old}$ with $\phi$. If $\alpha_{old}$ is full, auxiliary variables $j$ and $minDistance$ are initialized on lines $8$ and $9$. Subsequently, the loop in lines $10-19$ is performed for each solution in $\alpha_{new}$. In line $11$, the variable $sumDistances_{\phi_i}$ is initialized. Then, the sum of Hamming distances between each solution $\phi_i \in \alpha_{new}$ and all other solutions in $\alpha_{new}$ is computed and stored in $sumDistances_{\phi_i}$ (lines $12$ to $15$). If $sumDistances_{\phi_i}$ is smaller than the smallest known distance, variables $j$ and $minDistance$ are updated in lines $17$ and $18$. After the loop, the solution with the smallest Hamming distance is removed from the archive in line $21$. Finally, $\alpha_{new}$ is returned in line $23$. It is noteworthy that Algorithm \ref{alg:1} is easily adapted to JDAA by modifying the function in line $13$.

\setlength{\algomargin}{2em}
\begin{algorithm}[!ht]
    \KwIn{$\phi, \alpha_{old}$}
    \KwOut{$\alpha_{new}$}
    {
        \ForEach{$\phi_i \in \alpha_{old}$}{
            \If{$\phi \prec \phi_i$}{
                $\alpha_{old}$ $\leftarrow$ $\alpha_{old}$ - $\phi_i$
               
            }
        }
        $\alpha_{new}$ $\leftarrow$ $\alpha_{old} \cup \phi$
        
        \If{$archiveIsFull(\alpha_{old})$}{
                
            $k$ $\leftarrow$ $-1$

            $minDistance$ $\leftarrow$ $\infty$
                
            \ForEach{$\phi_i \in \alpha_{new}$}{
                $sumDistances_{\phi_i}$ $\leftarrow$ $0$
                
                \ForEach{$\phi_j \in \alpha_{new} - \phi_i$}{
                    $currDistance$ $\leftarrow$ $computeHammingDistance(\phi_i, \phi_j)$
                        
                    $sumDistances_{\phi_i}$ $\leftarrow$ $sumDistances_{\phi_i} + currDistance$
                }
                \If{$sumDistances_{\phi_i} < minDistance$}{
                    $k$ $\leftarrow$ $i$
                                    
                    $minDistance$ $\leftarrow$ $sumDistances_{\phi_i}$ 
                }
            }
            $\alpha_{new}$ $\leftarrow$ $\alpha_{new}$ - $\phi_k$
        }
        \Return $\alpha_{new}$
    }
    \caption{HDAA pseudocode. \label{alg:1}}
\end{algorithm}

\section{Numerical experiments} \label{s:experiments}

The bi-objective Travelling Salesman Problem~\cite{ribeiro2002}, which is the problem chosen to test the performance of the archiving algorithms, is described in Section~\ref{ss:bi-obj-tsp}. Following that, the Dominance-based multi-objective local search~\cite{liefooghe2012}, selected as the metaheuristic to generate the non-dominated solutions that are stored in the bounded archive, is detailed in Section~\ref{ss:pls}. Subsequently, the experiment protocol is outlined in Section~\ref{ss:protocol}. Finally, the computational tests conducted on the instance set defined in Section~\ref{ss:protocol} are presented in Section~\ref{ss:results}.

\subsection{Bi-Objective Travelling Salesman Problem} \label{ss:bi-obj-tsp}

The bi-objective Travelling Salesman Problem (Bi-obj TSP) is a multi-objective variant derived from the well-known Travelling Salesman Problem (TSP)~\cite{cook2007}. Initially introduced by~\cite{ribeiro2002}, it has since served as a benchmark for  evaluating the performance of various metaheuristics developed for MOOP~\cite{ribeiro2002,bouzid2022}. Given two distinct connected graphs $G_1=(N, E_1)$ and $G_2(N,E_2)$, where $N$ is the node-set and $E_1$ and $E_2$ are the edge-sets. Each edge $(i,j) \in E_1, E_2$ is associated with a non-negative cost $c_{ij}$. The Bi-obj TSP consists of finding, for each graph, the route with the smallest cost that visits all nodes $n \in N$ exactly once and returns to the initial node. Since the optimal routes for each graph are conflicting, the solution of the Bi-Obj TSP is represented by a Pareto front. On this front, no solution can be improved in one objective without adversely affecting the other.

\begin{figure}[!htbp]
\centering  
\subfigure[]  
{ 
\begin{tikzpicture}[>=latex',line join=bevel,][scale=.33,auto=left]
	\node (0) at (0bp,0bp) [draw,circle] {$0$};
	\node (1) at (60bp, 0bp) [draw,circle] {$1$};
	\node (2) at (0bp, 60bp) [draw,circle] {$2$};
	\node (3) at (60bp, 60bp) [draw,circle] {$3$};
				
	\draw [-] (0) -- node[below] {$4$} (1);
	\draw [-] (0) -- node[left] {$7$} (2);
	\draw [-] (0) -- node[above, near start]  {$9$} (3);
	\draw [-] (1) -- node[above, near start]  {$2$} (2);
	\draw [-] (1) -- node[right]  {$8$} (3);
	\draw [-] (2) -- node[above, sloped]  {$5$} (3);
\end{tikzpicture}%
} 
\hspace{3cm}
\subfigure[]  
{  
\begin{tikzpicture}[>=latex',line join=bevel,][scale=.33,auto=left]
	\node (0) at (0bp,0bp) [draw,circle] {$0$};
	\node (1) at (60bp, 0bp) [draw,circle] {$1$};
	\node (2) at (0bp, 60bp) [draw,circle] {$2$};
	\node (3) at (60bp, 60bp) [draw,circle] {$3$};
				
	\draw [-] (0) -- node[below] {$7$} (1);
	\draw [-] (0) -- node[left] {$5$} (2);
	\draw [-] (0) -- node[above, near start]  {$2$} (3);
	\draw [-] (1) -- node[above, near start]  {$6$} (2);
	\draw [-] (1) -- node[right]  {$4$} (3);
	\draw [-] (2) -- node[above, sloped]  {$9$} (3);
\end{tikzpicture}%
}
\caption{Example of instance of the bi-obj TSP. Each graph corresponds to one of the objectives.}
\label{fig:bi-obj-TSP}
\end{figure}
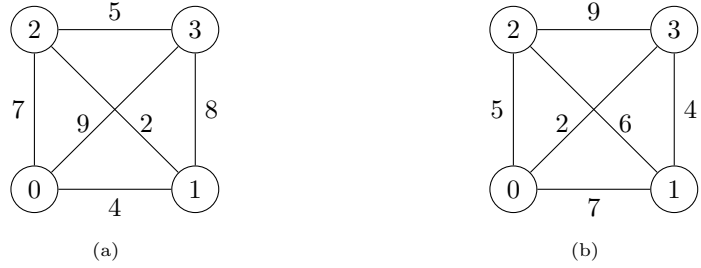

In Figure~\ref{fig:bi-obj-TSP}, an example of the Bi-obj TSP is presented. Graphs (a) and (b) within the figure depict the respective graphs to be optimized for each of the two objectives. For both graphs, the chosen initial node is set to $0$. For graph (a), the optimal solution is achieved by following the route $0 - 1 - 2 - 3 - 0$, incurring a total cost of $4 + 2 + 5 + 9 = 20$. Conversely, the optimal solution for graph (b) involves the path $0 - 3 - 1 - 2 - 0$, resulting in a total cost of $2 + 4 + 6 + 5 = 17$. It is noteworthy that the order of nodes in these optimal solutions differs, indicating the conflicting nature of the optimal solutions for both objectives of the Bi-obj TSP.

\subsection{Dominance-based multi-objective local search} \label{ss:pls}

The Dominance-based multi-objective local search (DMOLS) \cite{liefooghe2012} stands as a PLS metaheuristic designed for MOOP. It integrates the most advantageous features from various PLS approaches~\cite{knowles2000,talbi2001,angel2004}. DMOLS, which is characterized by a few parameters and several variants, has performed well across various MOOPs, as shown in the study~\cite{liefooghe2012}. In this research, a version of DMOLS integrated into the MH-Builder\footnote{MH-Builder, developed by the ORKAD team from the CRIStAL laboratory at the University of Lille, France, is a framework designed for developing adaptive metaheuristics for both single and multi-objective optimization problems.} software~\cite{mhbuilder} is employed to evaluate the performance of the archive algorithms outlined in Sections~\ref{s:obj-space} and~\ref{s:diversity}. 

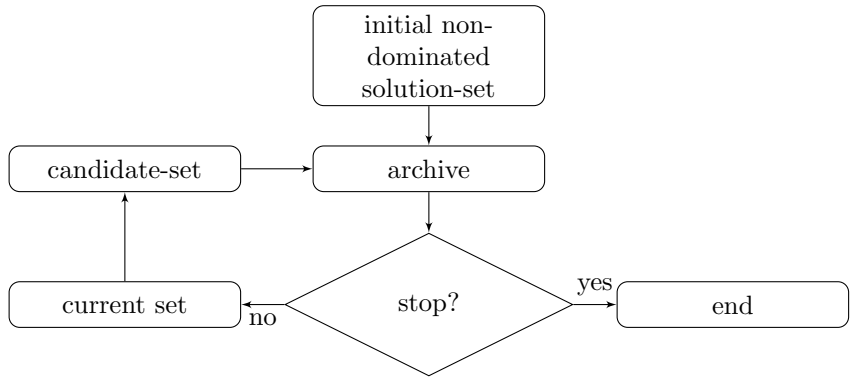
\begin{figure}[htbp]
	\centering
	\begin{tikzpicture}[node distance = 2cm, auto]

\tikzset{
  base/.style={draw, align=center, minimum height=4ex},
  proc/.style={base, rectangle, text width=8em},
  term/.style={proc, rounded corners},
  test/.style={base, diamond, aspect=2, text width=7em},
}

	% draw boxes
	\node [term] (init) {initial non-dominated solution-set};
    \node [term, below of=init, node distance=1.5cm] (archive) {archive};
    \node [test, below of=archive, node distance=1.8cm] (stopDMOLS) {stop?};
    \node [term, right of=stopDMOLS, node distance=4cm] (end) {end};
    \node [term, left of=stopDMOLS, node distance=4cm] (currs) {current set};
    \node [term, above of=currs, node distance=1.8cm] (cands) {candidate-set};
	
	% draw edges
	\path [line] (init) -- (archive);
	\path [line] (archive) -- (stopDMOLS);
	\path [line] (currs) --  (cands);
    \path [line] (cands) -- (archive);
    \path [line] (stopDMOLS.west) -- node{no} (currs.east);
    \path [line] (stopDMOLS.east) -- node{yes} (end.west);
	\end{tikzpicture}
	\caption{DMOLS flowchart. \label{fig:flowchart}}
\end{figure}

DMOLS operates following a sequence of steps. This metaheuristic starts with a set of nondominated solutions, usually generated by a greedy heuristic, which serves as an initialization of the archive.  Subsequently, DMOLS iterates through three steps until a specified stopping criterion is met, as shown in Figure \ref{fig:flowchart}. First, a subset of solutions from the archive is chosen to build the current set $C$. Following this, the neighborhood of $C$ is explored, creating a candidate set that contains new non-dominated solutions. Lastly, the archive is updated, incorporating new solutions identified within the candidate set and removing solutions that are dominated by the new ones. For a more comprehensive understanding of DMOLS and other PLS, readers are directed to~\cite{liefooghe2012}.

\subsection{Protocol} \label{ss:protocol}

The computational experiments are conducted on an Intel Xeon CPU W3520 with a 2.67 GHz clock speed, 24 cores, and 8 GB of RAM, running under the Linux operating system. All archiving algorithms are implemented in C++ and compiled with GNU g++ 7.5.0. These methods are incorporated into the MH-Builder~\cite{mhbuilder}. The proposed approaches were tested by solving the bi-obj TSP~\cite{ribeiro2002} with DMOLS~\cite{liefooghe2012} implemented into the MH-Builder and by limiting the maximum number of solutions in the archive.

The TSP instance set contains 45 complete graphs with up to 2000 nodes, based on instances introduced in the TSPLib~\cite{reinelt1991} and generated as suggested by~\cite{johnson2007}. The instance set is divided into three subsets consisting of 15 instances, each of which possesses a distinct graph structure and graphs of three different sizes (500, 1000, and 2000 nodes). In the first subset, referred to from now on as Random, the distance between each pair of nodes is chosen arbitrarily within a discrete uniform distribution $\mathscr{U}[0, 10^6]$. In the second subset, called Euclidean, all the nodes are randomly placed within a $10^6 \times 10^6$ plane employing a discrete uniform distribution $\mathscr{U}[1, 10^6]$ to generate their x- and y-coordinates. The last instance subset, named Cluster, is defined as follows. Firstly, a subset of nodes $N_c \subset N$ is arbitrarily put in a plane similar to the one defined for Euclidean instances. Next, the nodes in the node subset $N - N_c$ are distributed at random around the nodes in $N_c$. It is worth noting that the distance between a node in $N_c$ and another in $N - N_c$ belonging to the same cluster is at most $10^4$. The distance between two nodes in the Euclidean and Cluster subsets is determined by their Euclidean distance.

The creation of bi-obj TSP instances is accomplished by combining two distinct TSP instances of the same subset, each with the same number of nodes. As there are only three TSP instances of identical size within  each subset, there are three distinct ways to associate these instances. This results in 15 different bi-obj TSP instances for each subset. Therefore, the bi-obj TSP instance set also has 45 instances. Table~\ref{tab:instances} summarizes the main aspects of the bi-obj TSP instance set. 

\begin{table}[ht]
   \centering
   \setlength{\tabcolsep}{1.8em}
    \begin{tabular}{c|c}
			\multicolumn{1}{c|}{Characteristic} & Description \\ \hline 
			Graph sizes & 500, 1000, 2000 nodes \\
            Graph structure & Random, Euclidean, Cluster \\
            Instances per size & 3 \\
            Instances per type & 15 \\
            Total & 45
	\end{tabular}
    \caption{Main characteristics of the bi-obj TSP instance set. \label{tab:instances}}
\end{table}

\subsection{Results} \label{ss:results}

Tables~\ref{tab:demsar-hv-1} to~\ref{tab:demsar-igd-3} provide a comparative evaluation of solution diversity algorithms Random, AGA~\cite{knowles2002}, HA~\cite{knowles2003}, HDAA, and JDAA, conducted through a ranking analysis that employs the Dem\v{s}ar statistical test~\cite{demvsar2006}. The first two columns in each table indicate the number of nodes in each instance subset. Columns 3 through 7 display the average rankings assigned to each solution diversity algorithm based on their performance in a given instance size. The best algorithm for each instance-size subset is highlighted in bold. On average, HDAA emerged as the top-performing algorithm for both hypervolume and IGD+. The only exception to HDAA's dominance occurred when measuring the hypervolume for the Euclidean subset, where HA outperformed HDAA, signaling a nuanced outcome in this specific scenario. Additionally, it is noteworthy that, for instances with 500 nodes, HA consistently outperformed HDAA. However, this trend reversed for instances with 1000 and 2000 nodes, where HDAA demonstrated superior performance. This pattern suggests that HDAA generally performs better in larger instances, which are commonly encountered in real-world applications.

\begin{table}[htbp]
\centering
\footnotesize
\begin{tabular}{cc|c|c|c|c|c}
\multicolumn{ 2}{c|}{} & \multicolumn{ 5}{c}{Algorithm} \\
\multicolumn{ 2}{c|}{} & \multicolumn{1}{c|}{Random} & \multicolumn{1}{c|}{AGA} & \multicolumn{1}{c|}{HA} & \multicolumn{1}{c|}{HD} & \multicolumn{1}{c}{JD} \\ \hline
\multirow{3.1}{*}{Nodes}& 500 & 4.444 & 2.000 & \textbf{1.222} & 2.778 & 4.556 \\
& 1000 & 3.778 & 3.167 & 2.056 & \textbf{1.444} & 4.556 \\
& 2000 & 3.611 & 3.222 & 2.833 & \textbf{1.167} & 4.167 \\ \hline
\multicolumn{ 2}{c|}{Average} & 3.944 & 2.796 & 2.037 & \textbf{1.796} & 4.544 \\
\end{tabular}
\caption{Ranking of the solution diversity algorithms according to their hypervolume for the Cluster instance subset, using the statistical test introduced by~\cite{demvsar2006}. \label{tab:demsar-hv-1}}
\end{table}

\begin{table}[htbp]
\centering
\footnotesize
\begin{tabular}{cc|c|c|c|c|c}
\multicolumn{ 2}{c|}{} & \multicolumn{ 5}{c}{Algorithm} \\
\multicolumn{ 2}{c|}{} & \multicolumn{1}{c|}{Random} & \multicolumn{1}{c|}{AGA} & \multicolumn{1}{c|}{HA} & \multicolumn{1}{c|}{HD} & \multicolumn{1}{c}{JD} \\ \hline
\multirow{3.1}{*}{Nodes}& 500 & 4.500 & 1.889 & \textbf{1.333} & 2.778 & 4.500 \\
& 1000 & 3.722 & 3.056 & 2.278 & \textbf{1.556} & 4.389 \\
& 2000 & 4.556 & 1.889 & \textbf{1.111} & 3.000 & 4.444 \\ \hline
\multicolumn{ 2}{c|}{Average} & 4.259 & 2.278 & \textbf{1.574} & 2.444 & 4.444 \\
\end{tabular}
\caption{Ranking of the solution diversity algorithms according to their hypervolume for the Euclidean instance subset, using the statistical test introduced by~\cite{demvsar2006}. \label{tab:demsar-hv-2}}
\end{table}

\begin{table}[htbp]
\centering
\footnotesize
\begin{tabular}{cc|c|c|c|c|c}
\multicolumn{ 2}{c|}{} & \multicolumn{ 5}{c}{Algorithm} \\
\multicolumn{ 2}{c|}{} & \multicolumn{1}{c|}{Random} & \multicolumn{1}{c|}{AGA} & \multicolumn{1}{c|}{HA} & \multicolumn{1}{c|}{HD} & \multicolumn{1}{c}{JD} \\ \hline
\multirow{3.1}{*}{Nodes} & 500 & 4.333 & 1.667 & \textbf{1.389} & 2.944 & 4.667 \\
& 1000 & 3.778 & 2.889 & 2.111 & \textbf{1.556} & 4.667 \\
& 2000 & 3.667 & 3.111 & 2.778 & \textbf{1.444} & 4.000 \\ \hline
\multicolumn{ 2}{c|}{Average} & 3.926 & 2.556 & 2.093 & \textbf{1.981} & 4.444 \\
\end{tabular}
\caption{Ranking of the solution diversity algorithms according to their hypervolume for the Random instance subset, using the statistical test introduced by~\cite{demvsar2006}. \label{tab:demsar-hv-3}}
\end{table}

\begin{table}[htbp]
\centering
\footnotesize
\begin{tabular}{cc|c|c|c|c|c}
\multicolumn{ 2}{c|}{} & \multicolumn{ 5}{c}{Algorithm} \\
\multicolumn{ 2}{c|}{} & \multicolumn{1}{c|}{Random} & \multicolumn{1}{c|}{AGA} & \multicolumn{1}{c|}{HA} & \multicolumn{1}{c|}{HD} & \multicolumn{1}{c}{JD} \\ \hline
\multirow{3.1}{*}{Nodes}& 500 & 4.000 & 3.000 & 2.000 & \textbf{1.000} & 5.000 \\
& 1000 & 4.111 & 3.000 & 2.667 & \textbf{1.000} & 4.222 \\
& 2000 & 3.611 & 3.222 & 2.778 & \textbf{1.000} & 4.389 \\ \hline
\multicolumn{ 2}{c|}{Average} & 3.907 & 3.074 & 2.481 & \textbf{1.000} & 4.537 \\
\end{tabular}
\caption{Ranking of the solution diversity algorithms according to their IGD+ for the Cluster instance subset, using the statistical test introduced by~\cite{demvsar2006}. \label{tab:demsar-igd-1}}
\end{table}

\begin{table}[htbp]
\centering
\footnotesize
\begin{tabular}{cc|c|c|c|c|c}
\multicolumn{ 2}{c|}{} & \multicolumn{ 5}{c}{Algorithm} \\
\multicolumn{ 2}{c|}{} & \multicolumn{1}{c|}{Random} & \multicolumn{1}{c|}{AGA} & \multicolumn{1}{c|}{HA} & \multicolumn{1}{c|}{HD} & \multicolumn{1}{c}{JD} \\ \hline
\multirow{3.1}{*}{Nodes}& 500 & 4.444 & 2.389 & 2.611 & \textbf{1.000} & 4.556 \\
& 1000 & 3.556 & 2.722 & 2.889 & \textbf{1.000} & 4.833 \\
& 2000 & 3.500 & 3.611 & 2.944 & \textbf{1.000} & 3.944 \\ \hline
\multicolumn{ 2}{c|}{Average} & 3.833 & 2.907 & 2.815 & \textbf{1.000} & 4.444 \\
\end{tabular}
\caption{Ranking of the solution diversity algorithms according to their IGD+ for the Euclidean instance subset, using the statistical test introduced by~\cite{demvsar2006}. \label{tab:demsar-igd-2}}
\end{table}

\begin{table}[htbp]
\centering
\footnotesize
\begin{tabular}{cc|c|c|c|c|c}
\multicolumn{ 2}{c|}{} & \multicolumn{ 5}{c}{Algorithm} \\
\multicolumn{ 2}{c|}{} & \multicolumn{1}{c|}{Random} & \multicolumn{1}{c|}{AGA} & \multicolumn{1}{c|}{HA} & \multicolumn{1}{c|}{HD} & \multicolumn{1}{c}{JD} \\ \hline
\multirow{3.1}{*}{Nodes}& 500 & 4.056 & 2.556 & 2.444 & \textbf{1.000} & 4.944 \\
& 1000 & 4.000 & 3.056 & 2.389 & \textbf{1.000} & 4.556 \\
& 2000 & 3.611 & 3.167 & 3.167 & \textbf{1.000} & 4.056 \\ \hline
\multicolumn{ 2}{c|}{Average} & 3.889 & 2.926 & 2.667 & \textbf{1.000} & 4.519 \\
\end{tabular}
\caption{Ranking of the solution diversity algorithms according to their IGD+ for the Random instance subset, using the statistical test introduced by~\cite{demvsar2006}. \label{tab:demsar-igd-3}}
\end{table}

Tables~\ref{tab:3}-\ref{tab:11} provide a comparative analysis of the archive diversity algorithms for instances with 1000 nodes across the three subsets. Tables~\ref{tab:3}-\ref{tab:5} display results for archives with maximum capacities of 50, 100, and 200 solutions in the Cluster subset, while Tables~\ref{tab:6}-\ref{tab:8} and~\ref{tab:9}-\ref{tab:11} provide similar results for the Euclidean and Random subsets. For each instance, thirty runs are conducted for each algorithm, with distinct pseudorandom number generator seeds for each run. The maximum runtime per run was limited to a maximum of $600$ seconds. In these tables, columns 1 and 2 present the instance number and the solution diversity algorithm. Columns $3$-$5$ depict the average (Avg), median (Med), and standard deviation (Dev) for archive fullness, which measures the percentage of solutions at the end of each run relative to the archive's capacity. Similar data have been reported for the solution spread (Spread)~\cite{jiang2014}, the hypervolume (HV) of the Pareto front~\cite{auger2012}, and the Inverted Generational Distance (IGD+)~\cite{bezerra2017} in columns $6$-$8$, $9$-$11$, and $12$-$14$, respectively. The IGD+ metric quantifies the dissimilarity between the ideal Pareto front (the best solutions found by all algorithms) and the actual front. For each instance, the algorithms with the best spread, hypervolume, and IGD+ are highlighted in bold. Due to space limitations, only the results obtained with 1000 nodes are shown, as this is the point when HDAA emerges as the superior algorithm, both in HV and in IGD+. Also, it is worth mentioning that the results obtained for the 2000-node instances are similar to those with 1000 nodes. This consistency underlines the robustness and effectiveness of HDAA, particularly when dealing with larger problem sizes.
Further results for different instance sizes are available in Appendix \ref{sec:A1}.

The results derived from tables~\ref{tab:3}-\ref{tab:11} present notable insights into the performance of the algorithms Random, AGA, HA, HDAA, and JDAA. Notably, when the archive size is set at 50 and 100, all algorithms demonstrated an impressive utilization rate, using nearly 100\% of the archive. However, as the archive size increased to 200, all algorithms, except HDAA, utilized approximately 70\% of the archive, while HA stood out by employing, on average, around 95\% of the archive. Intriguingly, no algorithm exhibited clear superiority over others when considering solution spread. Nevertheless, HDAA and HA achieved superior hypervolumes, outperforming Random, AGA, and JDAA in this metric. In particular, HDAA obtained better results than all other methods in terms of IGD+, showcasing its prowess in solution diversity. These findings underscore the significance of exploring the solution space, with HDAA emerging as the most suitable algorithm for this purpose, emphasizing that the exploration of the solution space is, at the very least, as crucial as exploring the objective space.

\afterpage{%
\clearpage
\begin{landscape}
\centering
\footnotesize
\setcellgapes{1pt}
\makegapedcells
\setlength{\tabcolsep}{0.5em}
\begin{tabular}{c|c|ccc|ccc|ccc|ccc}
\multirow{2.1}{*}{Instance} & \multirow{2.1}{*}{Algorithm} & \multicolumn{3}{c|}{Fullness (\%)} & \multicolumn{3}{c|}{Spread} & \multicolumn{3}{c|}{HV} & \multicolumn{3}{c}{IGD+} \\
& & Avg & Med & Dev & Avg & Med & Dev & Avg & Med & Dev & Avg & Med & Dev \\ \hline
\multirow{5.1}{*}{1} & Random & 98.84 & 100.00 & 1.602 & \textbf{1.231} & \textbf{1.236} & \textbf{0.084} & 0.619 & 0.630 & 0.104 & 9.955E+06 & 9.639E+06 & 1.033E+06 \\
& AGA & 100.00 & 100.00 & 0.000 & 0.943 & 0.954 & 0.112 & 0.653 & 0.649 & 0.091 & 9.119E+06 & 9.043E+06 & 1.133E+06 \\
& HA & 100.00 & 100.00 & 0.000 & 0.791 & 0.801 & 0.121 & \textbf{0.662} & \textbf{0.666} & \textbf{0.087} & 9.012E+06 & 8.986E+06 & 1.039E+06 \\
& HDAA & 99.87 & 100.00 & 0.353 & 1.058 & 1.053 & 0.129 & 0.656 & 0.661 & 0.101 & \textbf{5.485E+06} & \textbf{5.642E+06} & \textbf{1.156E+06} \\
& JDAA & 99.81 & 100.00 & 0.390 & 1.186 & 1.180 & 0.132 & 0.533 & 0.549 & 0.095 & 1.129E+07 & 1.105E+07 & 1.639E+06 \\ \hline
\multirow{5.1}{*}{2} & Random & 99.94 & 100.00 & 0.177 & \textbf{1.248} & \textbf{1.287} & \textbf{0.116} & 0.628 & 0.637 & 0.100 & 1.087E+07 & 1.099E+07 & 1.497E+06 \\
\end{tabular}
\captionof{table}{Comparison of the solution diversity algorithms for the Cluster instance subset with 1000 nodes and an archive size of 50.\label{tab:3}}

\vspace{1cm}

\begin{tabular}{c|c|ccc|ccc|ccc|ccc}
\multirow{2.1}{*}{Instance} & \multirow{2.1}{*}{Algorithm} & \multicolumn{3}{c|}{Fullness (\%)} & \multicolumn{3}{c|}{Spread} & \multicolumn{3}{c|}{HV} & \multicolumn{3}{c}{IGD+} \\
& & Avg & Med & Dev & Avg & Med & Dev & Avg & Med & Dev & Avg & Med & Dev \\ \hline
\multirow{5.1}{*}{1} & Random & 98.03 & 100.00 & 4.329 & \textbf{1.131} & \textbf{1.137} & \textbf{0.099} & 0.646 & 0.658 & 0.100 & 8.146E+06 & 8.393E+06 & 1.252E+06 \\
& AGA & 99.29 & 100.00 & 1.669 & 1.021 & 1.013 & 0.107 & 0.670 & 0.691 & 0.092 & 7.861E+06 & 8.090E+06 & 1.172E+06 \\
& HA & 99.68 & 100.00 & 0.857 & 0.926 & 0.927 & 0.135 & \textbf{0.671} & \textbf{0.699} & \textbf{0.103} & 7.849E+06 & 8.090E+06 & 1.161E+06 \\
& HDAA & 97.58 & 100.00 & 4.195 & 1.052 & 1.040 & 0.096 & 0.675 & 0.696 & 0.092 & \textbf{5.941E+06} & \textbf{5.919E+06} & \textbf{1.207E+06} \\
& JDAA & 98.45 & 100.00 & 3.749 & 1.096 & 1.105 & 0.109 & 0.624 & 0.631 & 0.098 & 8.533E+06 & 8.530E+06 & 1.593E+06 \\ \hline
\multirow{5.1}{*}{2} & Random & 98.81 & 100.00 & 3.596 & \textbf{1.131} & \textbf{1.137} & \textbf{0.106} & 0.649 & 0.653 & 0.102 & 7.895E+06 & 7.946E+06 & 1.242E+06 \\
\end{tabular}
\captionof{table}{Comparison of the solution diversity algorithms for the Cluster instance subset with 1000 nodes and an archive size of 100. \label{tab:4}}
\end{landscape}
\clearpage% Flush page
}

\afterpage{%
\clearpage
\begin{landscape}
\centering
\footnotesize
\setcellgapes{1pt}
\makegapedcells
\setlength{\tabcolsep}{0.5em}
\begin{tabular}{c|c|ccc|ccc|ccc|ccc}
\multirow{2.1}{*}{Instance} & \multirow{2.1}{*}{Algorithm} & \multicolumn{3}{c|}{Fullness (\%)} & \multicolumn{3}{c|}{Spread} & \multicolumn{3}{c|}{HV} & \multicolumn{3}{c}{IGD+} \\
& & Avg & Med & Dev & Avg & Med & Dev & Avg & Med & Dev & Avg & Med & Dev \\ \hline
\multirow{5.1}{*}{1} & Random & 75.84 & 76.50 & 22.218 & 1.009 & 0.997 & 0.102 & 0.667 & 0.697 & 0.103 & 8.282E+06 & 8.551E+06 & 1.263E+06 \\
& AGA & 74.94 & 75.00 & 21.844 & \textbf{1.013} & \textbf{1.003} & \textbf{0.110} & 0.668 & 0.697 & 0.103 & 8.281E+06 & 8.524E+06 & 1.255E+06 \\
& HA & 75.13 & 76.00 & 22.600 & \textbf{1.013} & \textbf{0.996} & \textbf{0.111} & 0.668 & 0.697 & 0.102 & 8.268E+06 & 8.519E+06 & 1.260E+06 \\
& HDAA & 97.02 & 100.00 & 13.052 & 0.957 & 0.938 & 0.084 & \textbf{0.683} & \textbf{0.701} & \textbf{0.082} & \textbf{6.540E+06} & \textbf{6.744E+06} & \textbf{1.406E+06} \\
& JDAA & 75.39 & 75.00 & 21.916 & 1.010 & 0.995 & 0.110 & 0.667 & 0.697 & 0.103 & 8.275E+06 & 8.551E+06 & 1.257E+06 \\ \hline
\multirow{5.1}{*}{2} & Random & 72.92 & 70.00 & 20.969 & 1.039 & 1.040 & 0.123 & 0.662 & 0.662 & 0.100 & 7.769E+06 & 7.851E+06 & 1.174E+06 \\
\end{tabular}
\captionof{table}{Comparison of the solution diversity algorithms for the Cluster instance subset with 1000 nodes and an archive size of 200. \label{tab:5}}

\vspace{1cm}

\begin{tabular}{c|c|ccc|ccc|ccc|ccc}
\multirow{2.1}{*}{Instance} & \multirow{2.1}{*}{Algorithm} & \multicolumn{3}{c|}{Fullness (\%)} & \multicolumn{3}{c|}{Spread} & \multicolumn{3}{c|}{HV} & \multicolumn{3}{c}{IGD+} \\
& & Avg & Med & Dev & Avg & Med & Dev & Avg & Med & Dev & Avg & Med & Dev \\ \hline
\multirow{5.1}{*}{1} & Random & 98.77 & 100.00 & 1.430 & \textbf{1.149} & \textbf{1.117} & \textbf{0.112} & 0.635 & 0.651 & 0.083 & 9.126E+06 & 9.238E+06 & 1.240E+06 \\
& AGA & 99.94 & 100.00 & 0.177 & 0.945 & 0.974 & 0.111 & 0.685 & 0.708 & 0.079 & 8.504E+06 & 8.778E+06 & 1.309E+06 \\
& HA & 99.94 & 100.00 & 0.177 & 0.764 & 0.778 & 0.113 & \textbf{0.689} & \textbf{0.717} & \textbf{0.081} & 8.411E+06 & 8.475E+06 & 1.268E+06 \\
& HDAA & 99.42 & 100.00 & 1.590 & 1.008 & 1.002 & 0.097 & 0.659 & 0.687 & 0.076 & \textbf{5.366E+06} & \textbf{5.351E+06} & \textbf{1.295E+06} \\
& JDAA & 98.65 & 100.00 & 2.131 & 1.147 & 1.167 & 0.148 & 0.586 & 0.624 & 0.086 & 9.605E+06 & 9.894E+06 & 1.168E+06 \\ \hline
\multirow{5.1}{*}{2} & Random & 99.74 & 100.00 & 0.553 & \textbf{1.191} & \textbf{1.193} & \textbf{0.102} & 0.648 & 0.671 & 0.076 & 7.376E+06 & 7.394E+06 & 9.163E+05 \\
\end{tabular}
\captionof{table}{Comparison of the solution diversity algorithms for the Euclidean instance subset with 1000 nodes and an archive size of 50. \label{tab:6}}
\end{landscape}
\clearpage% Flush page
}

\afterpage{%
\clearpage
\begin{landscape}
\centering
\footnotesize
\setcellgapes{1pt}
\makegapedcells
\setlength{\tabcolsep}{0.5em}
\begin{tabular}{c|c|ccc|ccc|ccc|ccc}
\multirow{2.1}{*}{Instance} & \multirow{2.1}{*}{Algorithm} & \multicolumn{3}{c|}{Fullness (\%)} & \multicolumn{3}{c|}{Spread} & \multicolumn{3}{c|}{HV} & \multicolumn{3}{c}{IGD+} \\
& & Avg & Med & Dev & Avg & Med & Dev & Avg & Med & Dev & Avg & Med & Dev \\ \hline
\multirow{5.1}{*}{1} & Random & 97.45 & 100.00 & 4.196 & \textbf{1.015} & \textbf{1.007} & \textbf{0.118} & 0.660 & 0.681 & 0.077 & 7.563E+06 & 7.801E+06 & 1.369E+06 \\
& AGA & 99.61 & 100.00 & 1.127 & 0.951 & 0.953 & 0.097 & 0.675 & 0.696 & 0.075 & 7.500E+06 & 7.854E+06 & 1.334E+06 \\
& HA & 98.97 & 100.00 & 2.087 & 0.885 & 0.908 & 0.108 & 0.675 & 0.696 & 0.078 & 7.453E+06 & 7.677E+06 & 1.317E+06 \\
& HDAA & 99.16 & 100.00 & 2.329 & 0.964 & 0.955 & 0.103 & \textbf{0.682} & \textbf{0.693} & \textbf{0.080} & \textbf{5.757E+06} & \textbf{6.063E+06} & \textbf{1.469E+06} \\
& JDAA & 98.48 & 100.00 & 2.883 & 0.994 & 0.968 & 0.106 & 0.658 & 0.674 & 0.079 & 7.616E+06 & 7.841E+06 & 1.423E+06 \\ \hline
\multirow{5.1}{*}{2} & Random & 98.55 & 100.00 & 3.582 & \textbf{1.050} & \textbf{1.040} & \textbf{0.107} & 0.680 & 0.690 & 0.076 & 7.002E+08 & 7.004E+08 & 1.283E+06 \\
\end{tabular}
\captionof{table}{Comparison of the solution diversity algorithms for the Euclidean instance subset with 1000 nodes and an archive size of 100. \label{tab:7}}

\vspace{1cm}

\begin{tabular}{c|c|ccc|ccc|ccc|ccc}
\multirow{2.1}{*}{Instance} & \multirow{2.1}{*}{Algorithm} & \multicolumn{3}{c|}{Fullness (\%)} & \multicolumn{3}{c|}{Spread} & \multicolumn{3}{c|}{HV} & \multicolumn{3}{c}{IGD+} \\
& & Avg & Med & Dev & Avg & Med & Dev & Avg & Med & Dev & Avg & Med & Dev \\ \hline
\multirow{5.1}{*}{1} & Random & 66.08 & 66.00 & 15.668 & 0.949 & 0.954 & 0.112 & 0.677 & 0.689 & 0.078 & 7.361E+06 & 7.668E+06 & 1.295E+06 \\
& AGA & 66.27 & 67.00 & 15.709 & \textbf{0.950} & \textbf{0.950} & \textbf{0.111} & 0.677 & 0.695 & 0.079 & 7.361E+06 & 7.672E+06 & 1.299E+06 \\
& HA & 66.34 & 66.50 & 15.064 & 0.949 & 0.956 & 0.110 & 0.675 & 0.694 & 0.078 & 7.373E+06 & 7.668E+06 & 1.305E+06 \\
& HDAA & 93.77 & 98.00 & 14.927 & 0.860 & 0.869 & 0.094 & \textbf{0.705} & \textbf{0.716} & \textbf{0.075} & \textbf{5.806E+06} & \textbf{6.167E+06} & \textbf{1.542E+06} \\
& JDAA & 66.32 & 67.00 & 15.124 & 0.947 & 0.947 & 0.110 & 0.676 & 0.694 & 0.078 & 7.374E+06 & 7.672E+06 & 1.305E+06 \\ \hline
\multirow{5.1}{*}{2} & Random & 69.45 & 69.50 & 21.008 & 0.956 & 0.947 & 0.103 & 0.689 & 0.701 & 0.079 & 5.122E+06 & 5.057E+06 & 8.629E+05 \\
\end{tabular}
\captionof{table}{Comparison of the solution diversity algorithms for the Euclidean instance subset with 1000 nodes and an archive size of 200. \label{tab:8}}
\end{landscape}
\clearpage% Flush page
}

\afterpage{%
\clearpage
\begin{landscape}
\centering
\footnotesize
\setcellgapes{1pt}
\makegapedcells
\setlength{\tabcolsep}{0.5em}
\begin{tabular}{c|c|ccc|ccc|ccc|ccc}
\multirow{2.1}{*}{Instance} & \multirow{2.1}{*}{Algorithm} & \multicolumn{3}{c|}{Fullness (\%)} & \multicolumn{3}{c|}{Spread} & \multicolumn{3}{c|}{HV} & \multicolumn{3}{c}{IGD+} \\
& & Avg & Med & Dev & Avg & Med & Dev & Avg & Med & Dev & Avg & Med & Dev \\ \hline
\multirow{5.1}{*}{1} & Random & 99.55 & 100.00 & 0.869 & \textbf{1.262} & \textbf{1.264} & \textbf{0.086} & 0.645 & 0.647 & 0.077 & 6.588E+08 & 6.594E+08 & 2.451E+06 \\
& AGA & 99.94 & 100.00 & 0.177 & 1.061 & 1.056 & 0.105 & \textbf{0.679} & \textbf{0.673} & \textbf{0.073} & 6.583E+08 & 6.586E+08 & 2.298E+06 \\
& HA & 100.00 & 100.00 & 0.000 & 0.901 & 0.906 & 0.131 & 0.685 & 0.686 & 0.074 & 6.582E+08 & 6.583E+08 & 2.286E+06 \\
& HDAA & 99.55 & 100.00 & 0.705 & 1.071 & 1.065 & 0.111 & 0.677 & 0.662 & 0.074 & \textbf{6.544E+08} & \textbf{6.547E+08} & \textbf{2.447E+06} \\
& JDAA & 100.00 & 100.00 & 0.000 & 1.238 & 1.220 & 0.156 & 0.576 & 0.592 & 0.096 & 6.591E+08 & 6.594E+08 & 2.876E+06 \\ \hline
\multirow{5.1}{*}{2} & Random & 99.68 & 100.00 & 0.883 & \textbf{1.262} & \textbf{1.260} & \textbf{0.090} & 0.640 & 0.650 & 0.085 & 8.933E+06 & 9.127E+06 & 1.133E+06 \\
\end{tabular}
\captionof{table}{Comparison of the solution diversity algorithms for the Random instance subset with 1000 nodes and an archive size of 50.\label{tab:9}}

\vspace{1cm}

\begin{tabular}{c|c|ccc|ccc|ccc|ccc}
\multirow{2.1}{*}{Instance} & \multirow{2.1}{*}{Algorithm} & \multicolumn{3}{c|}{Fullness (\%)} & \multicolumn{3}{c|}{Spread} & \multicolumn{3}{c|}{HV} & \multicolumn{3}{c}{IGD+} \\
& & Avg & Med & Dev & Avg & Med & Dev & Avg & Med & Dev & Avg & Med & Dev \\ \hline
\multirow{5.1}{*}{1} & Random & 98.94 & 100.00 & 3.222 & \textbf{1.201} & \textbf{1.198} & \textbf{0.094} & 0.678 & 0.670 & 0.073 & 7.472E+06 & 7.574E+06 & 1.353E+06 \\
& AGA & 98.39 & 100.00 & 3.149 & 1.102 & 1.084 & 0.098 & 0.687 & 0.682 & 0.076 & 7.309E+06 & 7.368E+06 & 1.287E+06 \\
& HA & 99.03 & 100.00 & 2.694 & 1.041 & 1.050 & 0.131 & 0.691 & 0.689 & 0.075 & 7.284E+06 & 7.384E+06 & 1.309E+06 \\
& HDAA & 99.03 & 100.00 & 3.167 & 1.090 & 1.087 & 0.098 & \textbf{0.692} & \textbf{0.675} & \textbf{0.072} & \textbf{5.598E+06} & \textbf{5.694E+06} & \textbf{1.293E+06} \\
& JDAA & 99.06 & 100.00 & 3.121 & 1.160 & 1.150 & 0.097 & 0.654 & 0.641 & 0.079 & 7.684E+06 & 7.564E+06 & 1.559E+06 \\ \hline
\multirow{5.1}{*}{2} & Random & 97.45 & 100.00 & 5.217 & \textbf{1.197} & \textbf{1.206} & \textbf{0.095} & 0.658 & 0.691 & 0.083 & 6.584E+08 & 6.582E+08 & 1.640E+06 \\
\end{tabular}
\captionof{table}{Comparison of the solution diversity algorithms for the Random instance subset with 1000 nodes and an archive size of 100. \label{tab:10}}
\end{landscape}
\clearpage% Flush page
}

\afterpage{%
\clearpage
\begin{landscape}
\centering
\footnotesize
\setcellgapes{1pt}
\makegapedcells
\setlength{\tabcolsep}{0.5em}
\begin{tabular}{c|c|ccc|ccc|ccc|ccc}
\multirow{2.1}{*}{Instance} & \multirow{2.1}{*}{Algorithm} & \multicolumn{3}{c|}{Fullness (\%)} & \multicolumn{3}{c|}{Spread} & \multicolumn{3}{c|}{HV} & \multicolumn{3}{c}{IGD+} \\
& & Avg & Med & Dev & Avg & Med & Dev & Avg & Med & Dev & Avg & Med & Dev \\ \hline
\multirow{5.1}{*}{1} & Random & 76.84 & 73.00 & 27.043 & \textbf{1.124} & \textbf{1.133} & \textbf{0.108} & 0.697 & 0.699 & 0.073 & 7.218E+06 & 7.153E+06 & 1.278E+06 \\
& AGA & 76.32 & 73.50 & 27.264 & \textbf{1.124} & \textbf{1.133} & \textbf{0.109} & 0.697 & 0.705 & 0.073 & 7.187E+06 & 7.197E+06 & 1.277E+06 \\
& HA & 76.71 & 71.50 & 26.740 & 1.121 & 1.132 & 0.111 & 0.698 & 0.705 & 0.073 & 7.183E+06 & 7.152E+06 & 1.280E+06 \\
& HDAA & 97.84 & 100.00 & 8.525 & 1.049 & 1.058 & 0.083 & \textbf{0.713} & \textbf{0.720} & \textbf{0.065} & \textbf{5.804E+06} & \textbf{5.843E+06} & \textbf{1.437E+06} \\
& JDAA & 76.60 & 73.50 & 27.064 & \textbf{1.124} & \textbf{1.129} & \textbf{0.108} & 0.697 & 0.705 & 0.073 & 7.189E+06 & 7.192E+06 & 1.282E+06 \\ \hline
\multirow{5.1}{*}{2} & Random & 74.60 & 70.50 & 23.184 & \textbf{1.092} & \textbf{1.073} & \textbf{0.085} & 0.669 & 0.699 & 0.081 & 6.413E+06 & 6.499E+06 & 1.088E+06 \\
\end{tabular}
\captionof{table}{Comparison of the solution diversity algorithms for the Random instance subset with 1000 nodes and an archive size of 200. \label{tab:11}}
\end{landscape}
\clearpage% Flush page
}

\section{Concluding remarks and perspectives} \label{s:conclusion}

This work investigates solution diversity algorithms within bounded archives for multi-objective metaheuristics, particularly addressing challenges related to exponential growth in non-dominated solutions and the focus on a subset of the Pareto Front. Two established solution diversity algorithms from the literature, AGA~\cite{knowles2002} and HA~\cite{knowles2003}, primarily concentrating on the objective space, were compared against three innovative methodologies introduced in this study. The novel approaches, HDAA and JDAA, prioritize the exploration of the solution space. HDAA consistently stands out as a superior method, demonstrating effectiveness in enhancing solution diversity, especially in larger instances. The research emphasizes the importance of exploring the solution space, and HDAA emerges as a promising avenue for improving the efficiency of metaheuristics developed for solving MOOP. The findings underscore the significance of HDAA in addressing challenges posed by solution diversity, offering a robust and effective approach for optimizing metaheuristics in complex problem scenarios.

Considering future avenues of research, it is recommended to submit HDAA and JDAA to extensive testing across a spectrum of diverse multi-objective problems. This broader application will provide valuable insights into the generalizability and adaptability of these algorithms. Additionally, it is recommended to employ those solution diversity algorithms with other metaheuristics such as the NSGA-II and the MOEA/D. Evaluating the proposed algorithms with different metaheuristic frameworks offers a comprehensive understanding of their performance under varied optimization methodologies. Lastly, it is suggested for further advancement in the development of new algorithms designed to explore the solution space and enhance the solution diversity. Such innovations can contribute to a more robust and versatile set of tools for addressing multi-objective optimization challenges.

\section*{Data availability}

The datasets used in the course of the present study can be obtained by reaching out to the corresponding author upon request.

% --- Begin appendix ---
\appendix
\renewcommand{\thesection}{Appendix \Alph{section}}

\section{Extended results}\label{sec:A1}

Tables~\ref{tab:17}-\ref{tab:34} provide a comparative analysis of the archive diversity algorithms for instances with 500 and 2000 nodes across the three instance subsets defined in Section \ref{ss:protocol}. Tables~\ref{tab:17}-\ref{tab:22} display results for archives with maximum capacities of 50, 100, and 200 solutions in the Cluster subset, while Tables~\ref{tab:23}-\ref{tab:28} and~\ref{tab:29}-\ref{tab:34} provide similar results for the Euclidean and Random subsets. For each instance, thirty runs are conducted for each algorithm, with distinct pseudorandom number generator seeds for each run. The maximum runtime per run was limited to a maximum of $300$ seconds for instances with 500 nodes and $1200$ seconds for instances with 2000 nodes. In these tables, columns 1 and 2 present the instance number and the solution diversity algorithm. Columns $3$-$5$ depict the average (Avg), median (Med), and standard deviation (Dev) for archive fullness, which measures the percentage of solutions at the end of each run relative to the archive's capacity. Similar data have been reported for the solution spread, the hypervolume (HV) of the Pareto front, and the Inverted Generational Distance (IGD+) in columns $6$-$8$, $9$-$11$, and $12$-$14$, respectively. For each instance, the algorithms with the best spread, hypervolume, and IGD+ are highlighted in bold.

\afterpage{%
\clearpage
\begin{landscape}
\centering
\footnotesize
\setcellgapes{1pt}
\makegapedcells
\setlength{\tabcolsep}{0.5em}

%=== TABELA 17 ===%
\begin{tabular}{c|c|ccc|ccc|ccc|ccc}
\multirow{2}{*}{Instance} & \multirow{2}{*}{Algorithm}
& \multicolumn{3}{c|}{Fullness (\%)}
& \multicolumn{3}{c|}{Spread}
& \multicolumn{3}{c|}{HV}
& \multicolumn{3}{c}{IGD+} \\
& & Avg & Med & Dev & Avg & Med & Dev & Avg & Med & Dev & Avg & Med & Dev \\ \hline
\multirow{5}{*}{1} & Random & 98.71 & 100.00 & 2.163 & \textbf{1.239} & \textbf{1.253} & \textbf{0.098} & 0.599 & 0.604 & 0.069 & 2.073E+07 & 2.044E+07 & 1.564E+06 \\
& AGA & 98.71 & 100.00 & 2.560 & 0.772 & 0.754 & 0.085 & 0.753 & 0.759 & 0.053 & 2.020E+07 & 2.038E+07 & 1.554E+06 \\
& HA & 100.00 & 100.00 & 0.000 & 0.487 & 0.489 & 0.085 & \textbf{0.792} & \textbf{0.802} & \textbf{0.045} & 1.755E+07 & 1.748E+07 & 1.483E+06 \\
& HDAA & 99.55 & 100.00 & 1.237 & 0.989 & 0.979 & 0.138 & 0.693 & 0.703 & 0.072 & \textbf{4.591E+06} & \textbf{4.809E+06} & \textbf{1.209E+06} \\
& JDAA & 99.48 & 100.00 & 1.413 & 1.124 & 1.128 & 0.137 & 0.556 & 0.555 & 0.100 & 2.108E+07 & 2.116E+07 & 1.297E+06 \\ \hline
\multirow{5}{*}{2} & Random & 99.03 & 100.00 & 1.882 & \textbf{1.270} & \textbf{1.266} & \textbf{0.127} & 0.610 & 0.617 & 0.072 & 2.162E+07 & 2.153E+07 & 1.393E+06 \\
\end{tabular}
\captionof{table}{Comparison of the solution diversity algorithms for the Cluster instance subset with 500 nodes and an archive size of 50.\label{tab:17}}

\vspace{1cm}

%=== TABELA 18 ===%
\begin{tabular}{c|c|ccc|ccc|ccc|ccc}
\multirow{2}{*}{Instance} & \multirow{2}{*}{Algorithm}
& \multicolumn{3}{c|}{Fullness (\%)}
& \multicolumn{3}{c|}{Spread}
& \multicolumn{3}{c|}{HV}
& \multicolumn{3}{c}{IGD+} \\
& & Avg & Med & Dev & Avg & Med & Dev & Avg & Med & Dev & Avg & Med & Dev \\ \hline
\multirow{5}{*}{1} & Random & 100.00 & 100.00 & 0.000 & 1.321 & 1.323 & 0.074 & 0.669 & 0.680 & 0.068 & 6.628E+06 & 6.428E+06 & 1.217E+06 \\
& AGA & 100.00 & 100.00 & 0.000 & 1.201 & 1.211 & 0.083 & 0.676 & 0.681 & 0.064 & 6.534E+06 & 6.293E+06 & 1.253E+06 \\
& HA & 100.00 & 100.00 & 0.000 & 1.162 & 1.159 & 0.088 & 0.678 & 0.683 & 0.064 & 6.495E+06 & 6.227E+06 & 1.233E+06 \\
& HDAA & 100.00 & 100.00 & 0.000 & 1.208 & 1.213 & 0.075 & \textbf{0.692} & \textbf{0.704} & \textbf{0.052} & \textbf{5.477E+06} & \textbf{5.342E+06} & \textbf{1.184E+06} \\
& JDAA & 100.00 & 100.00 & 0.000 & \textbf{1.378} & \textbf{1.366} & \textbf{0.094} & 0.591 & 0.589 & 0.089 & 7.681E+06 & 7.195E+06 & 1.578E+06 \\ \hline
\multirow{5}{*}{2} & Random & 100.00 & 100.00 & 0.000 & 1.280 & 1.279 & 0.087 & 0.650 & 0.663 & 0.120 & 8.888E+06 & 9.134E+06 & 2.102E+06 \\
\end{tabular}
\captionof{table}{Comparison of the solution diversity algorithms for the Cluster instance subset with 2000 nodes and an archive size of 50. \label{tab:18}}
\end{landscape}
\clearpage
}

\afterpage{%
\clearpage
\begin{landscape}
\centering
\footnotesize
\setcellgapes{1pt}
\makegapedcells
\setlength{\tabcolsep}{0.5em}

\begin{tabular}{c|c|ccc|ccc|ccc|ccc}
\multirow{2.1}{*}{Instance} & \multirow{2.1}{*}{Algorithm} & \multicolumn{3}{c|}{Fullness (\%)} & \multicolumn{3}{c|}{Spread} & \multicolumn{3}{c|}{HV} & \multicolumn{3}{c}{IGD+} \\
& & Avg & Med & Dev & Avg & Med & Dev & Avg & Med & Dev & Avg & Med & Dev \\ \hline
\multirow{5.1}{*}{1} & Random & 99.35 & 100.00 & 2.496 & \textbf{1.127} & \textbf{1.136} & \textbf{0.105} & 0.656 & 0.662 & 0.081 & 1.304E+07 & 1.317E+07 & 1.015E+06 \\
& AGA & 99.97 & 100.00 & 0.177 & 0.818 & 0.820 & 0.064 & 0.765 & 0.778 & 0.053 & 1.212E+07 & 1.222E+07 & 1.179E+06 \\
& HA & 99.81 & 100.00 & 0.737 & 0.528 & 0.498 & 0.093 & \textbf{0.776} & \textbf{0.770} & \textbf{0.047} & 1.159E+07 & 1.171E+07 & 1.209E+06 \\
& HDAA & 98.71 & 100.00 & 3.985 & 1.060 & 1.078 & 0.111 & 0.716 & 0.721 & 0.068 & \textbf{5.025E+06} & \textbf{5.159E+06} & \textbf{1.274E+06} \\
& JDAA & 98.55 & 100.00 & 5.186 & 1.075 & 1.082 & 0.110 & 0.652 & 0.663 & 0.077 & 1.313E+07 & 1.323E+07 & 1.149E+06 \\ \hline
\multirow{5.1}{*}{2} & Random & 98.00 & 100.00 & 6.107 & 1.103 & 1.127 & 0.107 & 0.660 & 0.659 & 0.059 & 1.303E+07 & 1.310E+07 & 1.149E+06 \\
\end{tabular}
\captionof{table}{Comparison of the solution diversity algorithms for the Cluster instance subset with 500 nodes and an archive size of 100. \label{tab:19}}

\vspace{1cm}

\begin{tabular}{c|c|ccc|ccc|ccc|ccc}
\multirow{2.1}{*}{Instance} & \multirow{2.1}{*}{Algorithm} & \multicolumn{3}{c|}{Fullness (\%)} & \multicolumn{3}{c|}{Spread} & \multicolumn{3}{c|}{HV} & \multicolumn{3}{c}{IGD+} \\
& & Avg & Med & Dev & Avg & Med & Dev & Avg & Med & Dev & Avg & Med & Dev \\ \hline
\multirow{5.1}{*}{1} & Random & 86.61 & 86.00 & 10.567 & 1.307 & 1.309 & 0.068 & 0.678 & 0.683 & 0.064 & 7.758E+06 & 7.772E+06 & 1.122E+06 \\
& AGA & 86.52 & 85.00 & 10.169 & 1.302 & 1.302 & 0.068 & 0.678 & 0.683 & 0.064 & 7.757E+06 & 7.770E+06 & 1.116E+06 \\
& HA & 86.71 & 86.00 & 10.675 & 1.302 & 1.306 & 0.070 & 0.678 & 0.683 & 0.064 & 7.754E+06 & 7.770E+06 & 1.118E+06 \\
& HDAA & 99.71 & 100.00 & 0.990 & 1.224 & 1.217 & 0.064 & \textbf{0.699} & \textbf{0.709} & \textbf{0.046} & \textbf{6.552E+06} & \textbf{6.470E+06} & \textbf{1.107E+06} \\
& JDAA & 86.68 & 86.00 & 10.517 & \textbf{1.310} & \textbf{1.307} & \textbf{0.069} & 0.678 & 0.683 & 0.063 & 7.788E+06 & 7.770E+06 & 1.147E+06 \\ \hline
\multirow{5.1}{*}{2} & Random & 86.03 & 87.00 & 10.363 & 1.251 & 1.260 & 0.090 & 0.659 & 0.667 & 0.118 & 8.730E+06 & 9.284E+06 & 1.995E+06 \\
\end{tabular}
\captionof{table}{Comparison of the solution diversity algorithms for the Cluster instance subset with 2000 nodes and an archive size of 100.\label{tab:20}}

\end{landscape}
\clearpage% Flush page
}

\afterpage{%
\clearpage
\begin{landscape}
\centering
\footnotesize
\setcellgapes{1pt}
\makegapedcells
\setlength{\tabcolsep}{0.5em}

\begin{tabular}{c|c|ccc|ccc|ccc|ccc}
\multirow{2.1}{*}{Instance} & \multirow{2.1}{*}{Algorithm} & \multicolumn{3}{c|}{Fullness (\%)} & \multicolumn{3}{c|}{Spread} & \multicolumn{3}{c|}{HV} & \multicolumn{3}{c}{IGD+} \\
& & Avg & Med & Dev & Avg & Med & Dev & Avg & Med & Dev & Avg & Med & Dev \\ \hline
\multirow{5.1}{*}{1} & Random & 96.00 & 100.00 & 10.106 & 0.939 & 0.917 & 0.081 & 0.698 & 0.711 & 0.062 & 8.237E+06 & 8.442E+06 & 1.084E+06 \\
& AGA & 98.42 & 100.00 & 7.633 & 0.820 & 0.792 & 0.089 & 0.754 & 0.746 & 0.058 & 8.039E+06 & 8.316E+06 & 1.146E+06 \\
& HA & 98.76 & 100.00 & 5.494 & 0.672 & 0.628 & 0.112 & \textbf{0.763} & \textbf{0.773} & \textbf{0.055} & 7.845E+06 & 7.842E+06 & 1.052E+06 \\
& HDAA & 98.45 & 100.00 & 7.172 & \textbf{0.982} & \textbf{0.962} & \textbf{0.087} & 0.753 & 0.751 & 0.055 & \textbf{4.407E+06} & \textbf{4.664E+06} & \textbf{1.093E+06} \\
& JDAA & 97.52 & 100.00 & 10.222 & 0.922 & 0.906 & 0.074 & 0.713 & 0.710 & 0.054 & 8.489E+06 & 8.661E+06 & 1.138E+06 \\ \hline
\multirow{5.1}{*}{2} & Random & 95.82 & 100.00 & 12.734 & \textbf{1.027} & \textbf{1.041} & \textbf{0.088} & 0.707 & 0.717 & 0.064 & 8.846E+06 & 9.203E+06 & 1.123E+06 \\
\end{tabular}
\captionof{table}{Comparison of the solution diversity algorithms for the Cluster instance subset with 500 nodes and an archive size of 200.\label{tab:21}}

\vspace{1cm}

\begin{tabular}{c|c|ccc|ccc|ccc|ccc}
\multirow{2.1}{*}{Instance} & \multirow{2.1}{*}{Algorithm} & \multicolumn{3}{c|}{Fullness (\%)} & \multicolumn{3}{c|}{Spread} & \multicolumn{3}{c|}{HV} & \multicolumn{3}{c}{IGD+} \\
& & Avg & Med & Dev & Avg & Med & Dev & Avg & Med & Dev & Avg & Med & Dev \\ \hline
\multirow{5.1}{*}{1} & Random & 44.15 & 43.00 & 14.396 & \textbf{1.305} & \textbf{1.303} & \textbf{0.067} & 0.678 & 0.683 & 0.064 & 7.812E+06 & 7.626E+06 & 1.055E+06 \\
& AGA & 44.48 & 43.00 & 14.475 & 1.304 & 1.303 & 0.066 & 0.678 & 0.683 & 0.064 & 7.812E+06 & 7.626E+06 & 1.055E+06 \\
& HA & 44.55 & 44.00 & 14.414 & 1.304 & 1.304 & 0.067 & 0.678 & 0.683 & 0.064 & 7.813E+06 & 7.626E+06 & 1.052E+06 \\
& HDAA & 84.37 & 83.50 & 20.420 & 1.114 & 1.105 & 0.069 & \textbf{0.706} & \textbf{0.714} & \textbf{0.046} & \textbf{6.481E+06} & \textbf{6.342E+06} & \textbf{9.973E+05} \\
& JDAA & 44.35 & 43.00 & 14.461 & 1.304 & 1.303 & 0.067 & 0.678 & 0.683 & 0.064 & 7.812E+06 & 7.626E+06 & 1.052E+06 \\ \hline
\multirow{5.1}{*}{2} & Random & 43.44 & 43.50 & 10.983 & 1.251 & 1.261 & 0.087 & 0.659 & 0.667 & 0.118 & 8.814E+06 & 9.439E+06 & 2.047E+06 \\
\end{tabular}
\captionof{table}{Comparison of the solution diversity algorithms for the Cluster instance subset with 2000 nodes and an archive size of 200. \label{tab:22}}
\end{landscape}
\clearpage% Flush page
}

\afterpage{%
\clearpage
\begin{landscape}
\centering
\footnotesize
\setcellgapes{1pt}
\makegapedcells
\setlength{\tabcolsep}{0.5em}
\begin{tabular}{c|c|ccc|ccc|ccc|ccc}
\multirow{2.1}{*}{Instance} & \multirow{2.1}{*}{Algorithm} & \multicolumn{3}{c|}{Fullness (\%)} & \multicolumn{3}{c|}{Spread} & \multicolumn{3}{c|}{HV} & \multicolumn{3}{c}{IGD+} \\
& & Avg & Med & Dev & Avg & Med & Dev & Avg & Med & Dev & Avg & Med & Dev \\ \hline
\multirow{5.1}{*}{1} & Random & 98.52 & 100.00 & 4.064 & \textbf{1.147} & \textbf{1.151} & \textbf{0.097} & 0.593 & 0.604 & 0.058 & 1.582E+07 & 1.575E+07 & 1.054E+06 \\
& AGA & 99.87 & 100.00 & 0.246 & 0.772 & 0.785 & 0.108 & 0.692 & 0.690 & 0.032 & 1.425E+07 & 1.429E+07 & 8.658E+05 \\
& HA & 99.94 & 100.00 & 0.177 & 0.403 & 0.393 & 0.071 & \textbf{0.726} & \textbf{0.716} & \textbf{0.046} & 1.412E+07 & 1.440E+07 & 7.572E+05 \\
& HDAA & 100.00 & 100.00 & 0.000 & 0.915 & 0.898 & 0.130 & 0.648 & 0.661 & 0.054 & \textbf{3.123E+06} & \textbf{3.163E+06} & \textbf{8.531E+05} \\
& HDAA & 99.81 & 100.00 & 0.530 & 1.076 & 1.097 & 0.160 & 0.547 & 0.549 & 0.068 & 1.587E+07 & 1.581E+07 & 9.385E+05 \\
\hline
\multirow{5.1}{*}{2} & Random & 99.29 & 100.00 & 1.357 & \textbf{1.129} & \textbf{1.132} & \textbf{0.140} & 0.561 & 0.562 & 0.071 & 1.666E+07 & 1.702E+07 & 1.191E+06 \\
\end{tabular}
\captionof{table}{Comparison of the solution diversity algorithms for the Euclidean instance subset with 500 nodes and an archive size of 50.\label{tab:23}}

\vspace{1cm}

\begin{tabular}{c|c|ccc|ccc|ccc|ccc}
\multirow{2.1}{*}{Instance} & \multirow{2.1}{*}{Algorithm} & \multicolumn{3}{c|}{Fullness (\%)} & \multicolumn{3}{c|}{Spread} & \multicolumn{3}{c|}{HV} & \multicolumn{3}{c}{IGD+} \\
& & Avg & Med & Dev & Avg & Med & Dev & Avg & Med & Dev & Avg & Med & Dev \\ \hline
\multirow{5.1}{*}{1} & Random & 99.94 & 100.00 & 0.177 & 1.271 & 1.276 & 0.116 & 0.624 & 0.637 & 0.101 & 6.518E+06 & 6.830E+06 & 1.599E+06 \\
& AGA & 99.81 & 100.00 & 0.530 & 1.181 & 1.152 & 0.108 & 0.629 & 0.634 & 0.096 & 6.412E+06 & 6.638E+06 & 1.607E+06 \\
& HA & 100.00 & 100.00 & 0.000 & 1.112 & 1.098 & 0.136 & 0.632 & 0.640 & 0.096 & 6.374E+06 & 6.589E+06 & 1.597E+06 \\
& HDAA & 99.68 & 100.00 & 0.723 & 1.167 & 1.161 & 0.115 & \textbf{0.650} & \textbf{0.651} & \textbf{0.085} & \textbf{5.449E+06} & \textbf{5.449E+06} & \textbf{1.364E+06} \\
& JDAA & 99.87 & 100.00 & 0.353 & \textbf{1.285} & \textbf{1.279} & \textbf{0.117} & 0.553 & 0.550 & 0.097 & 7.517E+06 & 7.830E+06 & 2.124E+06 \\
\hline
\multirow{5.1}{*}{2} & Random & 100.00 & 100.00 & 0.000 & 1.340 & 1.359 & 0.091 & 0.640 & 0.624 & 0.085 & 7.320E+06 & 7.548E+06 & 1.014E+06 \\
\end{tabular}
\captionof{table}{Comparison of the solution diversity algorithms for the Euclidean instance subset with 2000 nodes and an archive size of 50. \label{tab:24}}
\end{landscape}
\clearpage% Flush page
}

\afterpage{%
\clearpage
\begin{landscape}
\centering
\footnotesize
\setcellgapes{1pt}
\makegapedcells
\setlength{\tabcolsep}{0.5em}

\begin{tabular}{c|c|ccc|ccc|ccc|ccc}
\multirow{2.1}{*}{Instance} & \multirow{2.1}{*}{Algorithm} & \multicolumn{3}{c|}{Fullness (\%)} & \multicolumn{3}{c|}{Spread} & \multicolumn{3}{c|}{HV} & \multicolumn{3}{c}{IGD+} \\
& & Avg & Med & Dev & Avg & Med & Dev & Avg & Med & Dev & Avg & Med & Dev \\ \hline
\multirow{5.1}{*}{1} & Random & 98.55 & 100.00 & 3.991 & \textbf{0.963} & \textbf{0.970} & \textbf{0.118} & 0.655 & 0.651 & 0.058 & 9.620E+06 & 9.693E+06 & 9.637E+05 \\
& AGA & 99.81 & 100.00 & 0.737 & 0.734 & 0.725 & 0.056 & 0.721 & 0.725 & 0.043 & 8.858E+06 & 8.962E+06 & 9.473E+05 \\
& HA & 99.87 & 100.00 & 0.707 & 0.469 & 0.444 & 0.105 & \textbf{0.722} & \textbf{0.723} & \textbf{0.045} & 9.040E+06 & 9.327E+06 & 8.602E+05 \\
& HDAA & 99.29 & 100.00 & 3.540 & 0.962 & 0.957 & 0.071 & 0.687 & 0.674 & 0.054 & \textbf{3.124E+06} & \textbf{3.101E+06} & \textbf{9.514E+05} \\
& JDAA & 99.00 & 100.00 & 2.615 & 0.939 & 0.926 & 0.102 & 0.661 & 0.651 & 0.056 & 9.570E+06 & 9.714E+06 & 8.498E+05 \\ \hline
\multirow{5.1}{*}{2} & Random & 99.23 & 100.00 & 2.195 & 0.985 & 0.961 & 0.106 & 0.645 & 0.649 & 0.064 & 3.328E+08 & 3.331E+08 & 1.554E+06 \\
\end{tabular}
\captionof{table}{Comparison of the solution diversity algorithms for the Euclidean instance subset with 500 nodes and an archive size of 100. \label{tab:25}}

\vspace{1cm}

\begin{tabular}{c|c|ccc|ccc|ccc|ccc}
\multirow{2.1}{*}{Instance} & \multirow{2.1}{*}{Algorithm} & \multicolumn{3}{c|}{Fullness (\%)} & \multicolumn{3}{c|}{Spread} & \multicolumn{3}{c|}{HV} & \multicolumn{3}{c}{IGD+} \\
& & Avg & Med & Dev & Avg & Med & Dev & Avg & Med & Dev & Avg & Med & Dev \\ \hline
\multirow{5.1}{*}{1} & Random & 81.23 & 83.00 & 11.321 & 1.227 & 1.197 & 0.102 & 0.631 & 0.637 & 0.096 & 6.158E+06 & 6.311E+06 & 1.533E+06 \\
& AGA & 81.45 & 83.00 & 11.193 & 1.227 & 1.196 & 0.103 & 0.632 & 0.637 & 0.095 & 6.158E+06 & 6.311E+06 & 1.533E+06 \\
& HA & 81.55 & 83.00 & 10.942 & 1.227 & 1.194 & 0.104 & 0.631 & 0.637 & 0.096 & 6.157E+06 & 6.311E+06 & 1.533E+06 \\
& HDAA & 98.68 & 100.00 & 4.223 & 1.142 & 1.121 & 0.117 & \textbf{0.664} & \textbf{0.659} & \textbf{0.086} & \textbf{5.133E+06} & \textbf{5.137E+06} & \textbf{1.284E+06} \\
& JDAA & 81.77 & 83.00 & 10.838 & \textbf{1.228} & \textbf{1.200} & \textbf{0.103} & 0.631 & 0.637 & 0.096 & 6.159E+06 & 6.311E+06 & 1.534E+06 \\ \hline
\multirow{5.1}{*}{2} & Random & 87.00 & 91.00 & 11.022 & \textbf{1.298} & \textbf{1.294} & \textbf{0.096} & 0.665 & 0.680 & 0.088 & 1.431E+09 & 1.431E+09 & 1.803E+06 \\
\end{tabular}
\captionof{table}{Comparison of the solution diversity algorithms for the Euclidean instance subset with 2000 nodes and an archive size of 100.\label{tab:26}}
\end{landscape}
\clearpage% Flush page
}

\afterpage{%
\clearpage
\begin{landscape}
\centering
\footnotesize
\setcellgapes{1pt}
\makegapedcells
\setlength{\tabcolsep}{0.5em}

\begin{tabular}{c|c|ccc|ccc|ccc|ccc}
\multirow{2.1}{*}{Instance} & \multirow{2.1}{*}{Algorithm} & \multicolumn{3}{c|}{Fullness (\%)} & \multicolumn{3}{c|}{Spread} & \multicolumn{3}{c|}{HV} & \multicolumn{3}{c}{IGD+} \\
& & Avg & Med & Dev & Avg & Med & Dev & Avg & Med & Dev & Avg & Med & Dev \\ \hline
\multirow{5.1}{*}{1} & Random & 97.69 & 100.00 & 8.400 & 0.768 & 0.780 & 0.101 & 0.696 & 0.686 & 0.052 & 7.036E+06 & 7.238E+06 & 9.805E+05 \\
& AGA & 98.84 & 100.00 & 7.127 & 0.713 & 0.716 & 0.087 & \textbf{0.718} & \textbf{0.713} & \textbf{0.051} & 6.755E+06 & 7.024E+06 & 9.459E+05 \\
& HA & 98.95 & 100.00 & 5.006 & 0.618 & 0.594 & 0.104 & 0.717 & 0.709 & 0.049 & 6.853E+06 & 7.012E+06 & 9.095E+05 \\
& HDAA & 98.16 & 100.00 & 9.690 & \textbf{0.823} & \textbf{0.803} & \textbf{0.094} & 0.713 & 0.712 & 0.045 & \textbf{3.617E+06} & \textbf{3.783E+06} & \textbf{9.551E+05} \\
& JDAA & 96.66 & 100.00 & 10.107 & 0.768 & 0.770 & 0.090 & 0.697 & 0.684 & 0.052 & 6.973E+06 & 7.173E+06 & 1.009E+06 \\ \hline
\multirow{5.1}{*}{2} & Random & 97.05 & 99.00 & 10.630 & 0.801 & 0.792 & 0.093 & 0.694 & 0.694 & 0.059 & 7.676E+06 & 7.731E+06 & 1.162E+06 \\
\end{tabular}
\captionof{table}{Comparison of the solution diversity algorithms for the Euclidean instance subset with 500 nodes and an archive size of 200.\label{tab:27}}

\vspace{1cm}

\begin{tabular}{c|c|ccc|ccc|ccc|ccc}
\multirow{2.1}{*}{Instance} & \multirow{2.1}{*}{Algorithm} & \multicolumn{3}{c|}{Fullness (\%)} & \multicolumn{3}{c|}{Spread} & \multicolumn{3}{c|}{HV} & \multicolumn{3}{c}{IGD+} \\
& & Avg & Med & Dev & Avg & Med & Dev & Avg & Med & Dev & Avg & Med & Dev \\ \hline
\multirow{5.1}{*}{1} & Random & 40.97 & 41.50 & 11.592 & \textbf{1.229} & \textbf{1.199} & \textbf{0.103} & 0.632 & 0.637 & 0.096 & 7.020E+06 & 7.067E+06 & 1.512E+06 \\
& AGA & 41.10 & 41.50 & 11.366 & 1.228 & 1.198 & 0.102 & 0.631 & 0.637 & 0.096 & 7.020E+06 & 7.067E+06 & 1.511E+06 \\
& HA & 40.79 & 41.00 & 10.989 & 1.229 & 1.200 & 0.102 & 0.631 & 0.637 & 0.096 & 7.019E+06 & 7.067E+06 & 1.509E+06 \\
& HDAA & 72.60 & 72.50 & 25.132 & 1.058 & 1.015 & 0.138 & \textbf{0.673} & \textbf{0.660} & \textbf{0.085} & \textbf{5.941E+06} & \textbf{5.851E+06} & \textbf{1.252E+06} \\
& JDAA & 41.03 & 41.50 & 11.144 & 1.228 & 1.198 & 0.103 & 0.631 & 0.637 & 0.096 & 7.018E+06 & 7.067E+06 & 1.511E+06 \\
\hline
\multirow{5.1}{*}{2} & Random & 43.31 & 44.00 & 12.846 & 1.295 & 1.286 & 0.097 & 0.665 & 0.680 & 0.088 & 7.024E+06 & 7.185E+06 & 9.365E+05 \\
\end{tabular}

\captionof{table}{Comparison of the solution diversity algorithms for the Euclidean instance subset with 2000 nodes and an archive size of 200. \label{tab:28}}
\end{landscape}
\clearpage% Flush page
}

\afterpage{%
\clearpage
\begin{landscape}
\centering
\footnotesize
\setcellgapes{1pt}
\makegapedcells
\setlength{\tabcolsep}{0.5em}
\begin{tabular}{c|c|ccc|ccc|ccc|ccc}
\multirow{2.1}{*}{Instance} & \multirow{2.1}{*}{Algorithm} & \multicolumn{3}{c|}{Fullness (\%)} & \multicolumn{3}{c|}{Spread} & \multicolumn{3}{c|}{HV} & \multicolumn{3}{c}{IGD+} \\
& & Avg & Med & Dev & Avg & Med & Dev & Avg & Med & Dev & Avg & Med & Dev \\ \hline
\multirow{5.1}{*}{1} & Random & 97.87 & 100.00 & 4.095 & \textbf{1.261} & \textbf{1.273} & \textbf{0.092} & 0.563 & 0.591 & 0.103 & 2.973E+08 & 2.975E+08 & 2.077E+06 \\
& AGA & 100.00 & 100.00 & 0.000 & 0.821 & 0.840 & 0.059 & 0.716 & 0.718 & 0.043 & 2.969E+08 & 2.975E+08 & 2.290E+06 \\
& HA & 100.00 & 100.00 & 0.000 & 0.461 & 0.463 & 0.093 & \textbf{0.763} & \textbf{0.768} & \textbf{0.053} & 2.958E+08 & 2.963E+08 & 1.911E+06 \\
& HDAA & 99.81 & 100.00 & 0.530 & 0.951 & 0.963 & 0.102 & 0.676 & 0.678 & 0.062 & \textbf{2.815E+08} & \textbf{2.815E+08} & \textbf{2.020E+06} \\
& JDAA & 96.90 & 100.00 & 4.406 & 1.137 & 1.124 & 0.158 & 0.551 & 0.564 & 0.087 & 2.976E+08 & 2.978E+08 & 1.952E+06 \\ \hline
\multirow{5.1}{*}{2} & Random & 98.97 & 100.00 & 1.776 & \textbf{1.219} & \textbf{1.232} & \textbf{0.117} & 0.564 & 0.549 & 0.083 & 1.941E+07 & 1.936E+07 & 1.497E+06 \\
\end{tabular}
\captionof{table}{Comparison of the solution diversity algorithms for the Random instance subset with 500 nodes and an archive size of 50.\label{tab:29}}

\vspace{1cm}

\begin{tabular}{c|c|ccc|ccc|ccc|ccc}
\multirow{2.1}{*}{Instance} & \multirow{2.1}{*}{Algorithm} & \multicolumn{3}{c|}{Fullness (\%)} & \multicolumn{3}{c|}{Spread} & \multicolumn{3}{c|}{HV} & \multicolumn{3}{c}{IGD+} \\
& & Avg & Med & Dev & Avg & Med & Dev & Avg & Med & Dev & Avg & Med & Dev \\ \hline
\multirow{5.1}{*}{1} & Random & 100.00 & 100.00 & 0.000 & 1.363 & 1.373 & 0.084 & 0.669 & 0.670 & 0.103 & 1.350E+09 & 1.351E+09 & 2.544E+06 \\
& AGA & 100.00 & 100.00 & 0.000 & 1.297 & 1.324 & 0.094 & 0.678 & 0.672 & 0.092 & 1.350E+09 & 1.351E+09 & 2.544E+06 \\
& HA & 100.00 & 100.00 & 0.000 & 1.246 & 1.245 & 0.099 & 0.679 & 0.674 & 0.092 & 1.350E+09 & 1.351E+09 & 2.551E+06 \\
& HDAA & 99.48 & 100.00 & 0.802 & 1.273 & 1.283 & 0.083 & \textbf{0.681} & \textbf{0.677} & \textbf{0.087} & \textbf{1.349E+09} & \textbf{1.349E+09} & \textbf{2.440E+06} \\
& JDAA & 100.00 & 100.00 & 0.000 & \textbf{1.470} & \textbf{1.458} & \textbf{0.082} & 0.567 & 0.569 & 0.133 & 1.351E+09 & 1.352E+09 & 3.272E+06 \\ \hline
\multirow{5.1}{*}{2} & Random & 100.00 & 100.00 & 0.000 & 1.359 & 1.372 & 0.100 & 0.675 & 0.704 & 0.101 & 9.560E+06 & 9.411E+06 & 1.565E+06 \\
\end{tabular}
\captionof{table}{Comparison of the solution diversity algorithms for the Random instance subset with 2000 nodes and an archive size of 50. \label{tab:30}}
\end{landscape}
\clearpage% Flush page
}

\afterpage{%
\clearpage
\begin{landscape}
\centering
\footnotesize
\setcellgapes{1pt}
\makegapedcells
\setlength{\tabcolsep}{0.5em}
\begin{tabular}{c|c|ccc|ccc|ccc|ccc}
\multirow{2.1}{*}{Instance} & \multirow{2.1}{*}{Algorithm} & \multicolumn{3}{c|}{Fullness (\%)} & \multicolumn{3}{c|}{Spread} & \multicolumn{3}{c|}{HV} & \multicolumn{3}{c}{IGD+} \\
& & Avg & Med & Dev & Avg & Med & Dev & Avg & Med & Dev & Avg & Med & Dev \\ \hline
\multirow{5.1}{*}{1} & Random & 99.65 & 100.00 & 1.033 & \textbf{1.128} & \textbf{1.138} & \textbf{0.105} & 0.639 & 0.666 & 0.085 & 3.070E+08 & 3.070E+08 & 2.121E+06 \\
& AGA & 99.71 & 100.00 & 1.419 & 0.797 & 0.784 & 0.077 & 0.740 & 0.755 & 0.063 & 3.063E+08 & 3.066E+08 & 2.060E+06 \\
& HA & 100.00 & 100.00 & 0.000 & 0.556 & 0.530 & 0.109 & \textbf{0.741} & \textbf{0.757} & \textbf{0.065} & 3.061E+08 & 3.065E+08 & 2.180E+06 \\
& HDAA & 99.03 & 100.00 & 2.901 & 1.020 & 1.005 & 0.097 & 0.690 & 0.704 & 0.069 & \textbf{2.992E+08} & \textbf{2.992E+08} & \textbf{2.036E+06} \\
& JDAA & 98.71 & 100.00 & 3.205 & 1.028 & 1.004 & 0.114 & 0.646 & 0.642 & 0.083 & 3.070E+08 & 3.071E+08 & 2.147E+06 \\
\hline
\multirow{5.1}{*}{2} & Random & 98.23 & 100.00 & 4.419 & \textbf{1.105} & \textbf{1.106} & \textbf{0.117} & 0.630 & 0.627 & 0.077 & 3.061E+08 & 3.067E+08 & 2.111E+06 \\
\end{tabular}
\captionof{table}{Comparison of the solution diversity algorithms for the Random instance subset with 500 nodes and an archive size of 100. \label{tab:31}}

\vspace{1cm}

\begin{tabular}{c|c|ccc|ccc|ccc|ccc}
\multirow{2.1}{*}{Instance} & \multirow{2.1}{*}{Algorithm} & \multicolumn{3}{c|}{Fullness (\%)} & \multicolumn{3}{c|}{Spread} & \multicolumn{3}{c|}{HV} & \multicolumn{3}{c}{IGD+} \\
& & Avg & Med & Dev & Avg & Med & Dev & Avg & Med & Dev & Avg & Med & Dev \\ \hline
\multirow{5.1}{*}{1} & Random & 86.94 & 86.00 & 10.429 & \textbf{1.406} & \textbf{1.399} & \textbf{0.074} & 0.679 & 0.674 & 0.092 & 7.902E+06 & 7.940E+06 & 1.299E+06 \\
& AGA & 87.26 & 88.00 & 9.811 & 1.405 & 1.405 & 0.074 & 0.679 & 0.674 & 0.092 & 7.902E+06 & 7.940E+06 & 1.300E+06 \\
& HA & 87.26 & 86.00 & 10.296 & 1.404 & 1.404 & 0.077 & 0.679 & 0.674 & 0.092 & 7.902E+06 & 7.940E+06 & 1.299E+06 \\
& HDAA & 98.87 & 100.00 & 2.882 & 1.338 & 1.345 & 0.071 & \textbf{0.686} & \textbf{0.687} & \textbf{0.087} & \textbf{6.862E+06} & \textbf{6.931E+06} & \textbf{1.276E+06} \\
& JDAA & 86.77 & 89.00 & 10.295 & 1.402 & 1.399 & 0.086 & 0.673 & 0.674 & 0.103 & 7.965E+06 & 7.940E+06 & 1.303E+06 \\ \hline
\multirow{5.1}{*}{2} & Random & 83.45 & 83.00 & 8.088 & 1.383 & 1.393 & 0.076 & 0.679 & 0.707 & 0.099 & 1.348E+09 & 1.349E+09 & 3.508E+06 \\
\end{tabular}
\captionof{table}{Comparison of the solution diversity algorithms for the Random instance subset with 2000 nodes and an archive size of 100.\label{tab:32}}
\end{landscape}
\clearpage% Flush page
}

\afterpage{%
\clearpage
\begin{landscape}
\centering
\footnotesize
\setcellgapes{1pt}
\makegapedcells
\setlength{\tabcolsep}{0.5em}

\begin{tabular}{c|c|ccc|ccc|ccc|ccc}
\multirow{2.1}{*}{Instance} & \multirow{2.1}{*}{Algorithm} & \multicolumn{3}{c|}{Fullness (\%)} & \multicolumn{3}{c|}{Spread} & \multicolumn{3}{c|}{HV} & \multicolumn{3}{c}{IGD+} \\
& & Avg & Med & Dev & Avg & Med & Dev & Avg & Med & Dev & Avg & Med & Dev \\ \hline
\multirow{5.1}{*}{1} & Random & 96.66 & 100.00 & 9.756 & 0.917 & 0.899 & 0.106 & 0.694 & 0.698 & 0.070 & 9.816E+06 & 9.820E+06 & 1.481E+06 \\
& AGA & 98.47 & 100.00 & 4.486 & 0.796 & 0.784 & 0.087 & \textbf{0.728} & \textbf{0.750} & \textbf{0.071} & 9.444E+06 & 9.397E+06 & 1.505E+06 \\
& HA & 98.26 & 100.00 & 6.010 & 0.657 & 0.634 & 0.102 & 0.727 & 0.748 & 0.067 & 9.623E+06 & 9.723E+06 & 1.577E+06 \\
& HDAA & 97.32 & 100.00 & 13.224 & \textbf{0.957} & \textbf{0.921} & \textbf{0.110} & 0.726 & 0.738 & 0.064 & \textbf{6.076E+06} & \textbf{6.024E+06} & \textbf{1.498E+06} \\
& JDAA & 98.58 & 100.00 & 6.658 & 0.887 & 0.882 & 0.116 & 0.693 & 0.687 & 0.076 & 1.007E+07 & 1.008E+07 & 1.555E+06 \\ \hline
\multirow{5.1}{*}{2} & Random & 95.47 & 99.50 & 11.187 & 0.919 & 0.906 & 0.090 & 0.682 & 0.678 & 0.073 & 1.049E+07 & 1.100E+07 & 1.843E+06 \\
\end{tabular}
\captionof{table}{Comparison of the solution diversity algorithms for the Random instance subset with 500 nodes and an archive size of 200.\label{tab:33}}

\vspace{1cm}

\begin{tabular}{c|c|ccc|ccc|ccc|ccc}
\multirow{2.1}{*}{Instance} & \multirow{2.1}{*}{Algorithm} & \multicolumn{3}{c|}{Fullness (\%)} & \multicolumn{3}{c|}{Spread} & \multicolumn{3}{c|}{HV} & \multicolumn{3}{c}{IGD+} \\
& & Avg & Med & Dev & Avg & Med & Dev & Avg & Med & Dev & Avg & Med & Dev \\ \hline
\multirow{5.1}{*}{1} & Random & 43.94 & 44.00 & 10.694 & 1.405 & 1.401 & 0.073 & 0.679 & 0.674 & 0.092 & 8.168E+06 & 8.164E+06 & 1.280E+06 \\
& AGA & 43.95 & 44.50 & 11.263 &\textbf{1.406} & \textbf{1.404} & \textbf{0.075} & 0.679 & 0.674 & 0.092 & 8.168E+06 & 8.164E+06 & 1.280E+06 \\
& HA & 44.24 & 44.50 & 10.842 & \textbf{1.406} & \textbf{1.404} & \textbf{0.073} & 0.679 & 0.674 & 0.092 & 8.168E+06 & 8.164E+06 & 1.279E+06 \\
& HDAA & 84.77 & 84.00 & 24.564 & 1.254 & 1.238 & 0.092 & \textbf{0.693} & \textbf{0.688} & \textbf{0.085} & \textbf{7.029E+06} & \textbf{7.000E+06} & \textbf{1.239E+06} \\
& JDAA & 44.03 & 45.00 & 11.545 & 1.405 & 1.404 & 0.074 & 0.679 & 0.674 & 0.092 & 8.168E+06 & 8.164E+06 & 1.279E+06 \\ \hline
\multirow{5.1}{*}{2} & Random & 41.69 & 41.50 & 9.328 & 1.383 & 1.385 & 0.075 & 0.679 & 0.707 & 0.099 & 8.878E+06 & 8.570E+06 & 1.415E+06 \\
\end{tabular}
\captionof{table}{Comparison of the solution diversity algorithms for the Random instance subset with 2000 nodes and an archive size of 200. \label{tab:34}}
\end{landscape}
\clearpage% Flush page
}

\bibliographystyle{sn-mathphys-num}%
\bibliography{sn-bibliography}% common bib file

\end{document}